\documentclass[11pt]{article}

\usepackage[final]{acl}

\usepackage{times}
\usepackage{latexsym}
\usepackage{url}
\usepackage{orcidlink}
\usepackage[T1]{fontenc}

\usepackage[utf8]{inputenc}

\usepackage{microtype}

\usepackage{inconsolata}

\usepackage{graphicx}
\usepackage{booktabs}
\usepackage{amsmath}
\usepackage{adjustbox}
\usepackage{enumerate}
\usepackage{enumitem}
\usepackage{fvextra}

\usepackage[most]{tcolorbox}
\newtcolorbox{promptbox}[1]{colback=gray!6, colframe=gray!55, boxrule=0.5pt,
  arc=1.5mm, left=2mm, right=2mm, top=1.5mm, bottom=1.5mm,
  title=#1, fonttitle=\footnotesize\bfseries, coltitle=black!70,
  colbacktitle=gray!15}

\title{HistXLS: Cross-Lingual Text Summarization of Historical Documents}

\author{
  \textbf{Matthias Schöffel\textsuperscript{1}}\,\orcidlink{0009-0004-4844-5372},
  \textbf{Marco Zeulner\textsuperscript{2}}\,\orcidlink{0009-0002-9276-2469},\\
  \textbf{Matthias Aßenmacher\textsuperscript{2}}\,\orcidlink{0000-0003-2154-5774},
  \textbf{Esteban Garcés Arias\textsuperscript{3,4}}\,\orcidlink{0009-0001-0467-8930}
\\
\\
  \textsuperscript{1}Bavarian Academy of Sciences (BAdW),
  \textsuperscript{2}Department of Statistics, LMU Munich,\\
  \textsuperscript{3}MaiNLP lab, Center for Information and Language Processing, LMU Munich,\\
  \textsuperscript{4}Munich Center for Machine Learning (MCML)
\\
\\
  \small{\textbf{Correspondence:} \href{mailto:matthias.schoeffel@badw.de}{matthias.schoeffel@badw.de}
  }
}

\begin{document}

\maketitle

\begin{abstract}
An edition of approximately 2,600 medieval Latin charters issued by Emperor Frederick~II (1194--1250) requires
\emph{Kopfregesten}---standardized single-sentence German summaries following strict diplomatic rules. 
Writing them manually demands senior Latin scholars and remains a major bottleneck in the multi-decade project. 
We investigate the potential of large language models (LLMs) for this low-resource cross-lingual summarization task, comparing few-shot prompting, modular translation/summarization pipelines, and parameter-efficient end-to-end finetuning across encoder--decoder and decoder-only architectures ranging from 0.3B to 72B parameters. 
To assess quality beyond standard NLP metrics, we develop an evaluation framework that aggregates 14 metrics across five equally weighted categories into a composite score in $[0,1]$, using an LLM-as-a-Judge to verify the diplomatic rules that surface-level metrics miss. 
A finetuned Qwen2.5-72B achieves the best overall score, and a finetuned 8B model surpasses the strongest 70B few-shot result; nevertheless, guideline-explicit few-shot prompting remains a competitive training-free baseline for large models, but it fails for small ones. 
Modular pipelines perform substantially worse: first-stage errors propagate irreversibly, as the second stage never attends to the original source. 
The finetuned small models (3B--8B) run on consumer hardware, making them deployable in a resource-constrained workflow as draft-generation tools with expert review.
\end{abstract}

\section{Introduction}
\label{sec:intro}

\begin{figure}
    \centering
    \includegraphics[angle=0,width=0.25\textwidth]{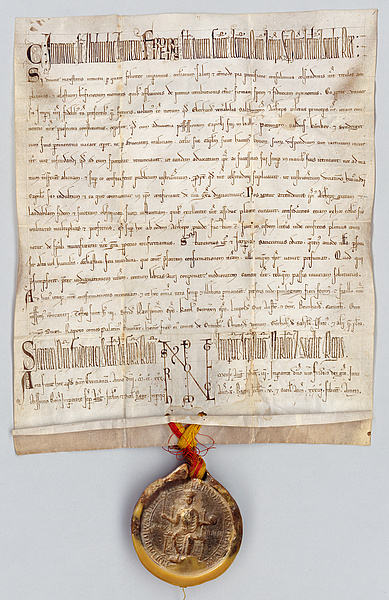}
    \caption{\textit{August 1230; Regesta Imperii V,1806; Bayerisches Hauptstaatsarchiv, Grafschaft Ortenburg Urk. 21} (Exemplary Charter)}
    \label{fig1}
\end{figure}

Medieval charters -- legal documents that offer crucial insights into the political, social, and economic structures of the Middle Ages -- constitute a significant part of humanity's still-unanalyzed textual heritage (see Figure \ref{fig1}). Among the vast corpora of unanalyzed historical texts are medieval charters of Emperor Frederick II of Hohenstaufen (1194--1250). The Bavarian Academy of Sciences (BAdW) is editing and publishing approximately 2,600 of these charters, scattered across European archives from Malta to St.~Petersburg, making this the largest international edition of medieval documents to date~\citep{friedrich_project}.

The editorial workflow relies on highly specialized Latin scholars who, after transcribing the medieval Latin manuscripts, create structured German summaries known as \textit{Kopfregesten}. These follow strict diplomatic conventions codified in the guidelines for the Edition of Sources on Regional History~\citep[\emph{Richtlinien für die Edition landesgeschichtlicher Quellen}; ][]{Heinemeyer2000}, which prescribe a specific structure capturing the essential legal content: the issuer, the dispositive legal action, and the recipient or beneficiary. This standardization ensures scholarly consistency and makes the charters' legal content broadly accessible. This traditional approach faces severe practical limitations. \textit{Kopfregest} creation is time-consuming and costly, requiring rare expertise in medieval Latin paleography, diplomatic conventions, and historical context. Only about 1,700 of the 2,600 charters have been published, with 14 years of Frederick II's reign still unprocessed -- and comparable editions for subsequent emperors will multiply this challenge. The scarcity of qualified experts thus constitutes a fundamental bottleneck.

Large language models (LLMs) offer new opportunities to automate parts of this workflow, particularly cross-lingual summarization (XLS) from medieval Latin to modern German. Unlike traditional NLP methods that require large parallel corpora, LLMs have shown strong capabilities in historical and low-resource languages \citep{Volk2024,Tekgurler2025}.

\paragraph {Problem Statement and Research Questions.} Applying LLMs to medieval Latin charter summarization poses challenges beyond standard XLS. First, training data is scarce: only about 1,700 charter--\textit{Kopfregest} pairs exist, far below the scale typical for neural summarization, motivating few-shot prompting and parameter-efficient finetuning (PEFT). Second, medieval Latin compounds this low-resource problem through strong orthographic variation across scribes, regions, and periods~\citep{Nowak2024}, regional language elements, and specialized diplomatic terminology; tokenizers
trained on modern languages fragment this vocabulary into character-level units~\citep{dorkin_tartunlp_2024}. Third, \textit{Kopfregesten} impose strict structural constraints: the output must be a single sentence following the issuer-verb-recipient structure and further diplomatic conventions, combining cross-lingual understanding, abstractive summarization, and constrained generation. Finally, standard metrics such as ROUGE or BLEU cannot capture adherence to these conventions, necessitating a specialized evaluation framework.

We address three aspects of our research question \textit{Can LLMs effectively automate the generation of German Kopfregesten from medieval Latin charters while maintaining historical fidelity and adherence to diplomatic conventions?}:

\begin{enumerate}[noitemsep]
    \item \textbf{Evaluation}: What combination of automated metrics and qualitative analysis captures historical accuracy and diplomatic correctness of generated \textit{Kopfregesten} best?
    
    \item \textbf{Architecture}: Do the theoretical advantages of encoder-decoder over decoder-only models for sequence-to-sequence tasks yield practical gains in this low-resource historical domain?
    
    \item \textbf{Training strategy}: Which approach offers the best trade-off between data efficiency, computational cost, and output quality?
\end{enumerate}

\paragraph{Contributions} We make the following contributions:

\begin{enumerate}[noitemsep]
     \item \textbf{Benchmarking LLMs on medieval Latin source text}: We present the first systematic evaluation of state-of-the-art models on medieval Latin to German summarization, enabled by an LLM-based extraction framework that recovers structured charter attributes more robustly than rule-based or regex approaches, informing model selection for similar digital humanities (DH) applications.
     

    
    
    \item \textbf{Comparison of XLS strategies}: We compare few-shot prompting to pipeline methods and PEFT across architectures, providing evidence-based guidance on aspects such as data/compute requirements and output quality under low-resource parallel data.

    \item \textbf{Specialized evaluation framework}: We propose a framework organizing 14 metrics into five equally weighted categories (\textbf{lexical}, \textbf{semantic}, \textbf{distributional}, \textbf{constraints}, and \textbf{efficiency}), each normalized to $[0,1]$; the composite score is the unweighted mean of the five category scores.
    
    \item \textbf{Practical implications}: We demonstrate that even small PEFT models achieve useful performance on consumer hardware, with direct relevance to DH digitization efforts, and provide concrete recommendations on model selection, training, and quality assurance.
    
    \item \textbf{Code, models, and data} are open-source: \url{https://github.com/tdslt-lmu/histxls}
\end{enumerate}

\section{Related Work}
\label{se:related_work}

\paragraph{Cross-lingual summarization.} XLS produces a summary in a target language different from the source document. Two paradigms recur: modular \emph{pipelines} that chain a translation and a summarization model---either
\textit{translate-then-summarize} \citep[TTS;][]{ouyang-etal-2019-robust,yao-etal-2015-phrase} or \textit{summarize-then-translate} \citep[STT;][]{orasan-chiorean-2008-evaluation,wan-etal-2010-cross}---and \emph{end-to-end} models trained directly on cross-lingually summarized pairs. Neural XLS was established through large-scale datasets and jointly trained architectures \citep{Zhu2019, Ladhak2020}, with XL-Sum \citep{Hasan2021} extending multilingual abstractive summarization to 44 languages. Surveys observe that pipelines are simple but suffer from error propagation, whereas end-to-end models avoid this at the cost of substantial parallel data \citep{Wang2022}. Our setting inverts the usual data assumption: the source is a low-resource \emph{historical} language with only ${\sim}1{,}700$ pairs, which rules out large-scale end-to-end training and makes the pipeline-vs-finetuning-vs-prompting comparison the central question rather than a mere implementation detail.

\paragraph{Cross-temporal summarization.} Closest to our setting are works that
summarize historical text into a modern language. \citet{peng-etal-2021-summarising}
introduce historical-to-modern summarization for historical German and
Chinese news and describe how lexical, syntactic, and stylistic change
complicates the task. CLCTS \citep{zhang-etal-2024-cross} extend this to
cross-lingual, cross-temporal summarization, compare pipeline, end-to-end, and
prompting approaches, and analyze the linguistic features of the historical sources.
Our setting differs from both in four respects. (i)~\emph{Time depth}: the sources
are Latin charters from 1199--1236, several centuries older than in both works.
(ii)~\emph{Content selection}: charters consist largely of fixed diplomatic formulae
(Appendix~\ref{a:dataset}), and the summary draws almost exclusively on the
\textit{Intitulatio} and \textit{Dispositio} (see the examples in
Appendix~\ref{a:examples}), so selection is determined by structure rather than by
salience. (iii)~\emph{Target norm}: \textit{Kopfregesten} are an editorial genre with
codified rules \citep{Heinemeyer2000}, which turns the task into constrained
generation. (iv)~\emph{References}: our targets were not collected or written for a benchmark but are the summaries of an ongoing scholarly edition, written by its editors under fixed conventions, so the task reproduces an existing editorial workflow.

\paragraph{NLP for historical and low-resource languages.} Historical languages compound data scarcity with strong orthographic variation \citep{arias-etal-2023-automatic, pavlopoulos-etal-2024-challenging, schoffel-etal-2025-modern, schoffel-garces-arias-2026-traditional, sarawgi-etal-2026-digitizing} and domain-specific vocabulary that modern tokenizers fragment \citep{Nowak2024, dorkin_tartunlp_2024}. Dedicated encoders such as Latin BERT \citep{Bamman2020} and the historically pretrained PhilTa/PhilBerta family \citep{Riemenschneider2023} improve coverage of classical and medieval Latin. More recently, general-purpose LLMs handle historical material with little or no task-specific training: \citet{Volk2024} report near-human cross-lingual summaries of 16th-century Latin letters, and \citet{Tekgurler2025} translate 18th-century Ottoman Turkish without explicit historical supervision. We extend this line to a task that is simultaneously cross-lingual, abstractive, and tightly constrained---the generation of diplomatically standardized \textit{Kopfregesten}. Work on handwritten text recognition for medieval Latin includes \citet{koch-etal-2023-tailored} for the digitization of record cards and \citet{peter_stotz_2021_4780947} who worked on handwritten Latin from the 16th century. Both works relate to ours through the shared language variety rather than through methodology.

\section{Data Source}
\label{sec:data}

\subsection{The Medieval Charter Dataset}
 
The whole dataset comprises 1,721 charters issued by Emperor Frederick~II, published by the {MGH}\footnote{\underline{M}onumenta \underline{G}ermaniae \underline{H}istorica: \url{https://www.dmgh.de/mgh_dd_f_ii_1/index.htm}} \citep{koch2002urkunden} in 10 Word documents of 153--927 pages each (Appendix~\ref{a:dataset}). Each charter entry combines editorial metadata added by modern historians with the diplomatic structure of the original Latin text. We extract two attributes per entry: the Latin charter text (model input) and the German \textit{Kopfregest} (reference output).

The diplomatic structure---the actual text of the charter---follows the standardized sections of medieval diplomatic practice (from \textit{Invocatio} to \textit{Apprecatio}), as described in detail in Appendix~\ref{a:dataset}.

\subsection{Extraction}
\label{subsec:extraction}
 
An initial regex-based extraction achieved only 84\% success: the documents lack fixed internal structure, and pattern matching cannot distinguish content by meaning (e.g., German summaries vs.\ German editorial annotations). We therefore redesigned the pipeline around {LLM}-based semantic extraction: documents are exported to {HTML}, segmented into per-charter blocks, automatically cleaned of editorial annotations (footnote markers and critical-apparatus paragraphs are removed by rule), and passed to Gemini~2.5~Flash with an engineered prompt enforcing verbatim transcription and edge-case handling. This raises extraction success to 97\%, with residual failures limited to extremely long charters or corrupted {HTML}. The extracted pairs were then manually checked by members of the edition project who provided the data. Full details of both approaches and the extraction prompt are given in Appendix~\ref{a:extraction}.

\begin{table}[ht]
\centering
\resizebox{0.4\textwidth}{!}{
\begin{tabular}{@{}lrrr@{}}
\toprule
                & \textbf{Mean} & \textbf{Median} & \textbf{Range} \\
\midrule
Latin input      & 2{,}738 & 2{,}064 & 52--16{,}818 \\
German output    &   296   &   274   & 56--2{,}914  \\
\bottomrule
\end{tabular}
}
\caption{Length statistics for both the Latin charters and the German summaries (measured in characters).}
\label{tab:lengths}
\end{table}

\begin{figure}[ht]
    \centering
    \includegraphics[width=0.49\textwidth]{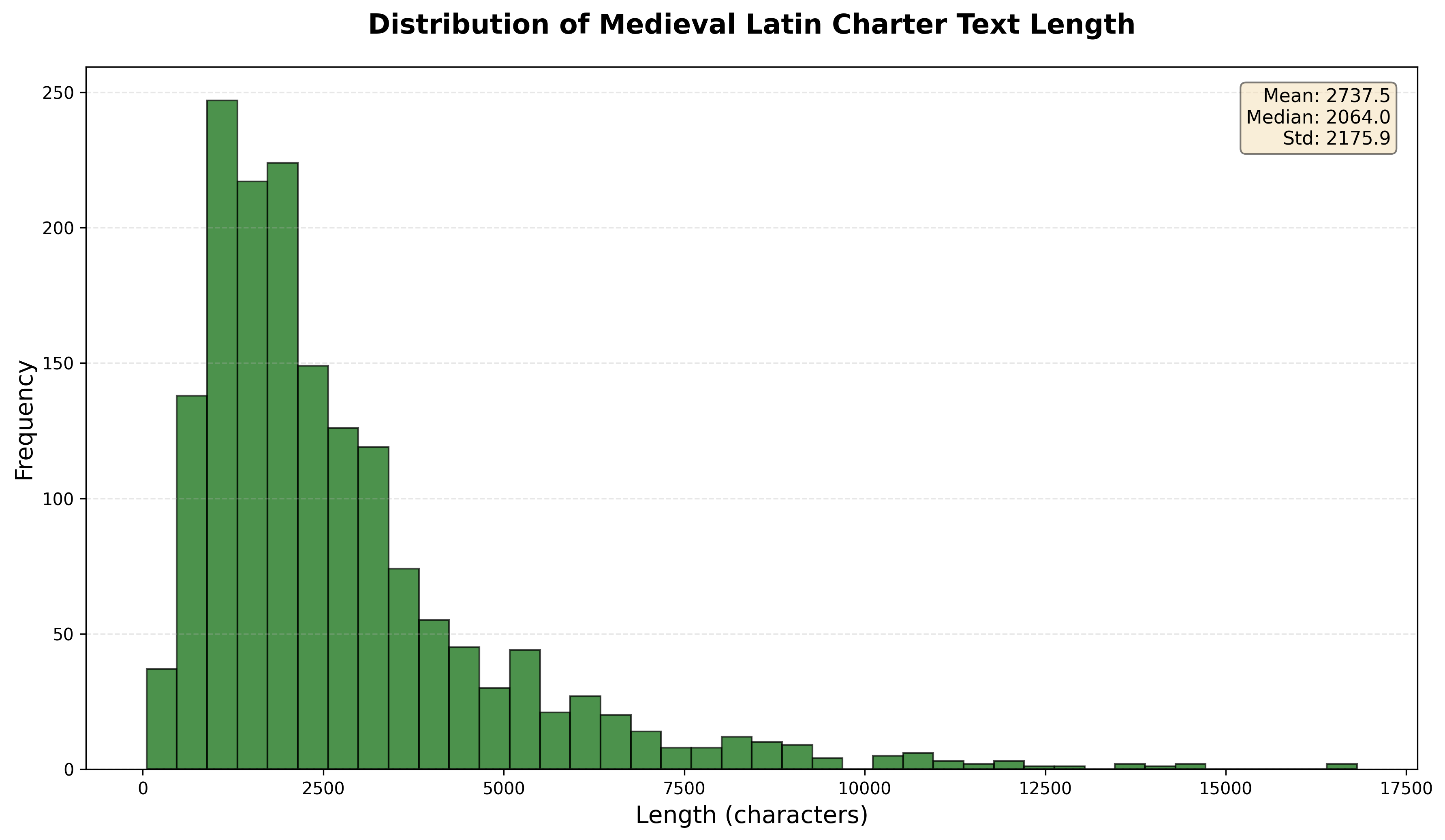}
    \includegraphics[width=0.49\textwidth]{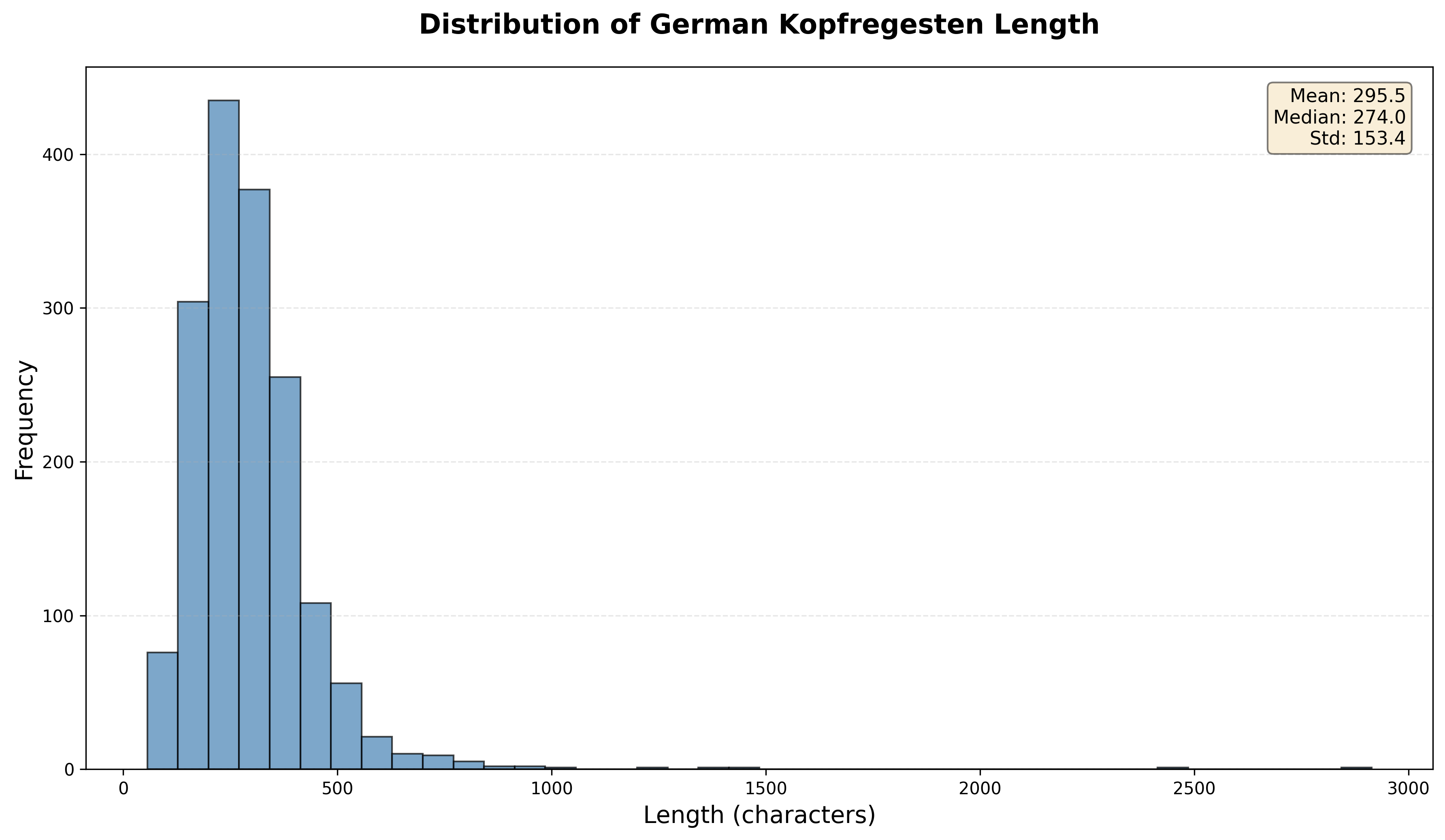}
    \caption{Length distributions (characters) of Latin charter texts (top)
    and German \textit{Kopfregesten} (bottom).}
    \label{fig:length_histograms}
\end{figure}

\subsection{Statistics and Split}
\label{subsec:statistics}
 
After preprocessing and quality control, the final dataset comprises 1,666 charter/summary pairs. Latin inputs average 2,738 characters (median 2,064; range 52--16,818) with a pronounced right skew, requiring careful context-window management during finetuning. German \textit{Kopfregesten} are far more constrained (mean 296, median 274 characters), reflecting the diplomatic requirement of concise single-sentence summaries; length distributions are shown in Table~\ref{tab:lengths} and Figure~\ref{fig:length_histograms}. Upper outliers among the summaries were extraction artifacts and were removed during quality control. On average, each summary is thus roughly a ninefold compression of its source text. The pairs are randomly partitioned 80--10--10 into 1,332 training, 167 validation, and 167 test samples. The split is used consistently: few-shot demonstrations are drawn exclusively from the training set, finetuned models are monitored on the validation set, and all approaches are evaluated on the held-out test set.

\section{Evaluation Framework}
\label{subsec:evaluation-framework}

Following a multicriteria perspective on text evaluation
\citep{garces-arias-etal-2025-towards}, we assess complementary
dimensions of summary quality. Reference-based metrics---BLEU \citep{Papineni2002}, ROUGE \citep{Lin2004}, BERTScore \citep{Zhang2020}, and MAUVE \citep{Pillutla2021}---capture lexical, semantic, and distributional fidelity but are blind to task-specific structural conventions. We therefore add an LLM-as-a-Judge as a deterministic binary checker of six explicit diplomatic rules, sidestepping the biases of open-ended scoring \citep{Zheng2023, Liu2023} while measuring rule adherence that surface metrics miss. None of the components verifies whether the facts of a summary match the charter (see Limitations).
 
\paragraph{Lexical.}
ROUGE-1/2/L capture unigram, bigram, and longest-common-subsequence overlap; SacreBLEU~\citep{Post2018} adds a precision-based perspective. Combining all four reduces the sensitivity of any single n-gram metric to decoding artifacts; the category score is their mean, scaled to $[0,1]$.
 
\paragraph{Semantic.}
{BERTScore} F1 is computed with the domain-specific \texttt{magistermilitum/\allowbreak bert\_medieval\_multilingual} model (layer 12), capturing semantic equivalence beyond surface wording. A cross-lingual variant (Cross-{BS}) between the generated German summary and the original Latin source directly measures cross-lingual fidelity---something no monolingual metric can capture. The category score is the mean of both F1 values.
 
\paragraph{Distributional.}
{MAUVE}~\citep{Pillutla2021}, computed with the same domain-specific embedding model, compares the full output distribution against the reference distribution, detecting corpus-level pathologies (mode collapse, systematic verbosity, stylistic drift) that pair-level metrics miss.
 
\paragraph{Efficiency.}
The length ratio (prediction/reference) is mapped to $[0,1]$ via 
\begin{equation*}
    S_{\text{Eff}} = \exp\!\left(-\left|1 - \text{Ratio}\right|\right),
    \label{eq:efficiency}
\end{equation*}
penalizing both verbosity and inappropriate brevity, which neither lexical nor semantic metrics capture directly.
 
\paragraph{Constraints ({LLM}-as-a-Judge).}
Qwen3-8B~\citep[4-bit, deterministic decoding;][]{qwen3technicalreport} checks each generated summary against six expert rules derived from the \textit{Kopfregest} conventions:

\begin{enumerate}[noitemsep]
    \item \textbf{Single sentence}: exactly one sentence (abbreviations such as \textit{z.B.} do not count as boundaries);
    \item \textbf{Issuer first}: the summary begins with the issuer's name;
    \item \textbf{Dispositive verb second}: a dispositive verb (e.g., \textit{best\"atigt}, \textit{schenkt}) immediately follows;
    \item \textbf{Recipient third}: the beneficiary is named immediately after the verb;
    \item \textbf{Modern names}: geographic and personal names appear in modern German form;
    \item \textbf{No forbidden content}: no threat clauses, witness lists, or dating formulas.
\end{enumerate}

Each rule is judged using a dedicated yes/no prompt; per-rule violation rates are inverted and averaged to form the category score.

\begin{figure*}[ht]
    \centering
    \includegraphics[width=0.85\textwidth]{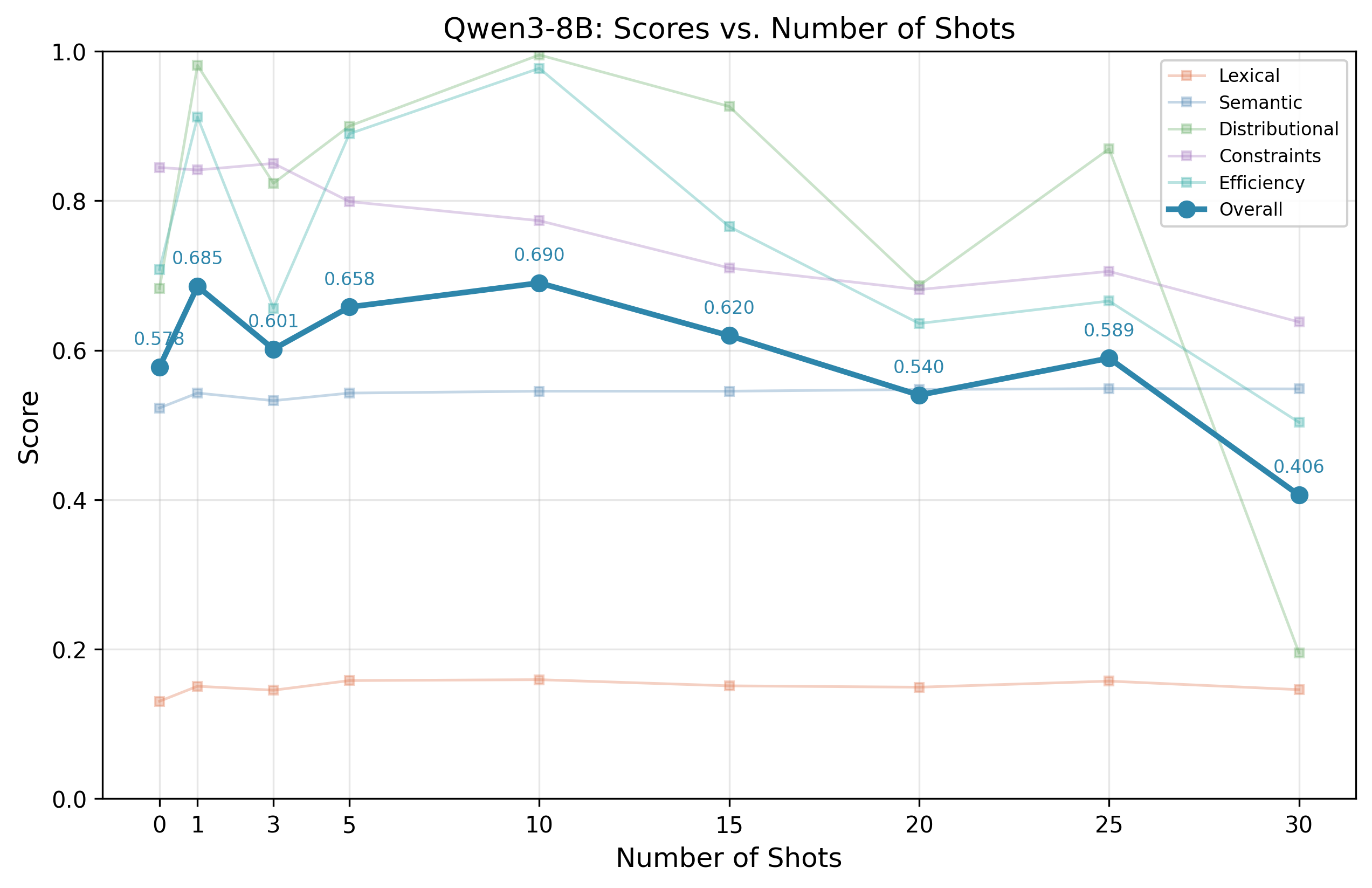}
    \caption{Effect of the number of demonstrations $n$ on \texttt{Qwen3-8B}
    (guideline-explicit prompt). Overall performance peaks at $n=10$; the
    Constraints score decreases monotonically with $n$.}
    \label{fig:fewshot_nshots_curve}
\end{figure*}
 
\paragraph{Composite score.} The overall score is the unweighted mean of the five category scores (see Table~\ref{tab:eval-categories}),
\begin{equation*}
    S_{\text{Overall}} = \tfrac{1}{5}\bigl(S_{\text{Lex}} + S_{\text{Sem}}
    + S_{\text{Dist}} + S_{\text{Cons}} + S_{\text{Eff}}\bigr),
\end{equation*}

avoiding researcher degrees of freedom in weighting. Averaging at the category level rather than over all 14 metrics prevents categories with many metrics (Lexical, Constraints) from dominating the score. Equal weighting is the standard default when no validated preference weights are available~\citep{Dawes1979,Nardo2008}; since we lack human quality judgments to fit weights against, we adopt it. 

\begin{table}[ht]
    \centering
\resizebox{0.47\textwidth}{!}{
    \begin{tabular}{p{1.5cm}p{2.5cm}p{2.8cm}}
        \toprule
        \textbf{Category} & \textbf{Metrics} & \textbf{Normalization} \\
        \midrule
        Lexical & {BLEU}, ROUGE-1, ROUGE-2, ROUGE-L & mean of 4 \\[2pt]
        Semantic & {BS} F1, Cross-{BS} F1 & mean of 2 \\[2pt]
        Distrib. & {MAUVE} & identity \\[2pt]
        Constraints & 6 rule violation rates VR (judge) & $\frac{1}{6}\sum_i (1-VR_i)$ \\[2pt]
        Efficiency & length ratio & $\exp(-|1-\text{Ratio}|)$ \\
        \bottomrule
    \end{tabular}
}
    \caption{The five evaluation categories with their respective metrics, and the normalization strategy to $[0,1]$.}
    \label{tab:eval-categories}
\end{table}
 
A sensitivity analysis (Appendix~\ref{a:weights}) shows that the ranking is robust to this choice: over 10{,}000 random weightings, the mean Kendall's $\tau$ with the reported ranking is $0.91$, and the top model is unchanged in 99\% of cases. Decomposition into category scores reveals \emph{where} a model excels: high Lexical but poor Constraints indicates mimicry of reference wording without absorbing the conventions; the converse indicates well-structured \textit{Kopfregesten} that merely differ in phrasing.
 
\section{Experiments}
\label{sec:experiments}
 
\subsection{Few-Shot Prompting}
\label{subsec:few-shot-prompting}
 
We test whether instruction-following LLMs can perform the full Latin-to-German {XLS} task via in-context learning~\citep[ICL;][]{OpenAI2020}, without parameter updates. This setting serves both as a training-free baseline and as an ablation of how far explicit prompt constraints alone can enforce diplomatic correctness. Since these constraints resemble instruction following~\citep{Ouyang2022}, only instruction-tuned chat models are considered: \texttt{Llama-3.3-70B-Instruct}~\citep{Grattafiori2024}, \texttt{Qwen2.5-72B-Instruct}~\citep{qwen25}, \texttt{Qwen3-8B}~\citep{qwen3technicalreport}, and \texttt{Qwen3-0.6B}.

\paragraph{Prompt design.} Because few-shot performance is sensitive to prompt wording and demonstration selection~\citep{Zhao2021}, we optimize the prompt in a controlled manner on \texttt{Qwen3-8B} with $n{=}10$ demonstrations drawn uniformly at random from the training set (fixed seed; see Appendix~\ref{a:fewshot}). A naive ``summarize the document'' prompt achieved $S_{\text{Overall}}=0.58$; encoding the diplomatic conventions as explicit enumerated rules (guideline-explicit prompt, Appendix~\ref{a:prompts}) improved this to $0.69$ ($+0.11$), driven mainly by Distributional/Efficiency gains. Tokenizer statistics (Appendix~\ref{a:tokenizer}) confirm a domain mismatch: 2.24 tokens per Latin word and a 0.15 single-character-token rate inflate effective sequence length.
 
\paragraph{Number of shots.} Sweeping $n \in \{0,1,3,5,10,15,20,25,30\}$, performance peaks at $n{=}10$; notably, the Constraints score decreases monotonically with $n$ (from $0.84$ at $n{=}0$ to $0.64$ at $n{=}30$): a larger demonstration context appears to dilute the explicit rule instructions rather than reinforce them (see Figure~\ref{fig:fewshot_nshots_curve}).

\paragraph{Results.} Table~\ref{tab:fewshot_overall_scores} reports all four models under the best configuration. The two $\sim$70B models are nearly tied (\texttt{Llama-3.3-70B-Instruct}: $0.712$; \texttt{Qwen2.5-72B}: $0.708$), with Llama ahead through marginally higher Lexical and Constraints scores. Lexical scores are low across the board but increase with scale. \texttt{Qwen3-0.6B} collapses, most visibly in its Distributional score of $0.027$ (vs.\ $\geq 0.995$ for all others), indicating outputs that diverge fundamentally from the reference distribution. Qualitative examples for the best model are shown in Appendix~\ref{a:examples} (Table~\ref{tab:fewshot_examples}).
 
\begin{table}[ht]
    \centering
    \resizebox{.47\textwidth}{!}{
    \begin{tabular}{lrrrrrr}
        \toprule \textbf{Model} & $S_{\text{Lex}}$ & $S_{\text{Sem}}$ & $S_{\text{Dist}}$ & $S_{\text{Cons}}$ & $S_{\text{Eff}}$ & $S_{\text{Overall}}$ \\
        \midrule \texttt{Llama-3.3-70B-Instruct} & \textbf{0.233} & \textbf{0.561}  & 0.997   & \textbf{0.846} & 0.925 & \underline{\textbf{0.712}}       \\
        \texttt{Qwen2.5-72B-Instruct}   & 0.217  & \textbf{0.561}  & \textbf{1.000}  & 0.830   & 0.934 & 0.708     \\
        \texttt{Qwen3-8B}     & 0.159   & 0.545   & 0.995  & 0.773 & \textbf{0.977}  &  0.690  \\
        \texttt{Qwen3-0.6B}   & 0.065  & 0.478  & 0.027  & 0.699   & 0.833  & 0.420    \\
        \bottomrule
    \end{tabular}
    }
    \caption{Few-shot results (guideline-explicit prompt, $n=10$ shots). Best score per column in bold.}
    \label{tab:fewshot_overall_scores}
\end{table}
 
\subsection{Pipeline Strategies}
\label{subsec:pipeline-strategies}
 
Modular pipelines decompose the task so that each stage can be handled by a specialized model: \textbf{{TTS}} (translate the Latin charter to German, then summarize) and \textbf{{STT}} (summarize within Latin, then translate the abstract). 
We evaluate five configurations of off-the-shelf encoder--decoder models---two {TTS} (\texttt{NLLB-200}~\citep{NLLBTeam2024} and
\texttt{MADLAD-400-3B}~\citep{Kudugunta2023}, each followed by \texttt{mT5-Small}~\citep{ml6Team2021}) and three {STT} (\texttt{LatinSummarizer}~\citep{Delaval2025}, \texttt{LongT5-Large}~\citep{Guo2022}, and \texttt{mT5-XLSum}~\citep{Hasan2021}, each followed by \texttt{NLLB-200}).
Large decoder-only models are excluded here, as using a 70B model as a single pipeline stage offers no theoretical benefit over its end-to-end use (Section~\ref{subsec:few-shot-prompting}). 
All models use beam search (4 beams); language-control and prefix details are given in Appendix~\ref{a:pipeline}.
 
\paragraph{Results.}
Table~\ref{tab:pipeline_scores_overall} reports all five configurations. The best, \texttt{MADLAD-400}~$\rightarrow$~\texttt{mT5-Small} ($0.511$), remains far below the few-shot baseline; \texttt{NLLB-200}-based {TTS} fails almost completely ($0.170$), and all {STT} configurations fail ($\leq 0.275$).
  
\begin{table}[ht]
    \centering
    \resizebox{.4\textwidth}{!}{
    \begin{tabular}{llr}
        \toprule \textbf{Dir.} & \textbf{Configuration} & $S_{\text{Overall}}$ \\
        \midrule
        TTS & \texttt{MADLAD-400 $\rightarrow$ mT5-Small}     & \underline{\textbf{0.511}} \\
        TTS & \texttt{NLLB-200 $\rightarrow$ mT5-Small}       & 0.170 \\
        \midrule
        STT & \texttt{LatinSummarizer $\rightarrow$ NLLB-200} & 0.275 \\
        STT & \texttt{mT5-XLSum $\rightarrow$ NLLB-200}       & 0.230 \\
        STT & \texttt{LongT5-Large $\rightarrow$ NLLB-200}    & 0.136 \\
        \bottomrule
    \end{tabular}
    }
    \caption{Overall pipeline results for both directions (full results in Appendix \ref{a:pipeline_full}). No configuration
    approaches the few-shot baseline. Best overall score in bold.}
    \label{tab:pipeline_scores_overall}
\end{table}

\paragraph{Failure modes.}
All failures trace to error propagation from the first stage. 
In {TTS}, \texttt{NLLB-200} exhibits syntactic echoing and character-level collapse on medieval Latin, while \texttt{MADLAD-400} produces grammatical but semantically distorted German (e.g., mistranslating \textit{curas} as \textit{Pferde}), leaving the summarizer without the key entities required for a valid \textit{Kopfregest}. 
In {STT}, no off-the-shelf model can summarize medieval Latin: \texttt{LatinSummarizer} emits sentinel tokens, \texttt{LongT5} degenerates into repetitive keyword lists, and \texttt{mT5-XLSum} hallucinates modern ecclesiastical content (e.g., ``Pope
Benedict~XVI''), which the translator then faithfully renders. 
Illustrative examples are given in Appendix~\ref{a:examples} (Table~\ref{tab:pipeline_tts_examples}).

\subsection{End-to-End Finetuning}
\label{subsec:end-to-end-finetuning}
 
Finetuning directly optimizes the model on paired charters and gold \textit{Kopfregesten}, addressing the two challenges prompting leaves open: domain mismatch with modern pretraining corpora, and consistent constraint adherence. It also promises resource efficiency: if small finetuned models match large few-shot models, the same quality becomes available at far lower inference cost.
 
\paragraph{Models.}
We finetune four encoder--decoder models---\texttt{PhilTa} \citep{Riemenschneider2023}, the only model pretrained on historical texts, \texttt{mT5-Small}/\texttt{mT5-Base}~\citep{Xue2021}, and \texttt{MADLAD-400} (3B machine-translation checkpoint, the same one used in the pipeline; Appendix~\ref{a:encdec_models}), enabling a direct pipeline-vs-finetuning comparison of the same model---and the four decoder-only models (see Section~\ref{subsec:few-shot-prompting}) as \emph{base} checkpoints, enabling a controlled per-model comparison between prompting and finetuning across two magnitudes.
 
\paragraph{Setup.} All models are trained with {LoRA}/{QLoRA} \citep{Hu2021,Dettmers2023}, adapter placement is described in Appendices~\ref{a:finetune} and~\ref{a:compute}. Hyperparameters were selected via random search on \texttt{Qwen3-0.6B} (10 epochs per trial, objective $S_{\text{Overall}}$; search space and best configuration in Appendix~\ref{a:hpo}) and applied unchanged to all models---a pragmatic compromise that ensures a clean comparison under a fixed training recipe. The spread across search runs was small ($\Delta S_{\text{Overall}}=0.03$), indicating robustness to the searched range. Training dynamics show clear overfitting after epoch~5 (validation loss minimum $0.866$ at step~300 for \texttt{Qwen2.5-72B}, rising to $1.110$ by step~600; Appendix~\ref{a:loss}); all results use the best checkpoint by validation loss.
 
\paragraph{Results.} Table~\ref{tab:e2e_results_all} reports all eight finetuned models. \texttt{Qwen2.5-72B} achieves the best overall score ($0.744$). The best encoder--decoder model, \texttt{MADLAD-400} ($0.707$), narrows the gap considerably but still trails the decoder-only family. Category radars per family and qualitative examples are provided in Appendices~\ref{a:radars} and~\ref{a:examples} (Table~\ref{tab:e2e_examples}).
 
\begin{table}[ht]
    \centering
    \resizebox{.49\textwidth}{!}{
    \begin{tabular}{lrrrrrr}
        \toprule \textbf{Model} & $S_{\text{Lex}}$ & $S_{\text{Sem}}$ & $S_{\text{Dist}}$ & $S_{\text{Cons}}$ & $S_{\text{Eff}}$ & $S_{\text{Overall}}$ \\
        \midrule \texttt{MADLAD-400}  & 0.314 & \textit{0.584} & 0.990 & 0.820 & 0.827 & 0.707 \\
        \texttt{mT5-Base}             & 0.282 & 0.575 & 0.863 & 0.813 & 0.767 & 0.660 \\
        \texttt{mT5-Small}            & 0.201 & 0.550 & 0.762 & 0.799 & 0.785 & 0.619 \\
        \texttt{PhilTa}               & 0.207 & 0.520 & 0.751 & 0.786 & 0.731 & 0.599 \\
        \midrule \texttt{Qwen2.5-72B} & \textit{0.332} & \textbf{0.588} & \textbf{0.999} & \textbf{0.843} & \textbf{0.956} & \underline{\textbf{0.744}} \\
        \texttt{Llama-3.3-70B}        & \textbf{0.334} & \textbf{0.588} & 0.994 & 0.831 & 0.874 & \textit{0.724} \\
        \texttt{Qwen3-8B}             & 0.290 & 0.577 & \textit{0.995} & \textit{0.833} & \textit{0.891} & 0.717 \\
        \texttt{Qwen3-0.6B}           & 0.240 & 0.557 & 0.971 & 0.827 & 0.791 & 0.677 \\
        \bottomrule
    \end{tabular}
    }
    \caption{End-to-end finetuning results. Encoder--decoder (top) and decoder-only (bottom) models, each sorted by descending overall score. Best score per column in bold, second-best in italics.}
    \label{tab:e2e_results_all}
\end{table}
 
\section{Final Results and Discussion}
\label{sec:results}
 
\subsection{Cross-Approach Comparison}
\label{subsec:performance-comparison}
 
Table~\ref{tab:cross_approach_ranking} ranks all 17 model--approach combinations. Finetuned models occupy the top three positions, with the best few-shot models interleaved at ranks 4--5 and all pipelines in the bottom third. Three observations stand out. First, the best finetuned model (\texttt{Qwen2.5-72B}, $0.744$) beats the best few-shot model (\texttt{Llama-3.3-70B-Instruct}, $0.712$) by only $0.032$ points. Second, a finetuned 8B model ($0.717$) already surpasses the best few-shot result of models with $\sim$9$\times$ more parameters. Third, the best pipeline ($0.511$) falls $0.20$--$0.23$ points below both end-to-end paradigms. 
  
\begin{table}[ht]
    \centering
    \resizebox{.49\textwidth}{!}{
    \begin{tabular}{clllr}
        \toprule
        \textbf{Rank} & \textbf{Model} & \textbf{Approach} & \textbf{Arch.} & $S_{\text{Overall}}$ \\
        \midrule
        1  & \texttt{Qwen2.5-72B}            & Finetuning & Dec.\ & \textbf{0.744} \\
        2  & \texttt{Llama-3.3-70B}           & Finetuning & Dec.\ & 0.724 \\
        3  & \texttt{Qwen3-8B}               & Finetuning & Dec.\ & 0.717 \\
        4  & \texttt{Llama-3.3-70B-Instruct}  & Few-shot   & Dec.\ & 0.712 \\
        5  & \texttt{Qwen2.5-72B-Instruct}    & Few-shot   & Dec.\ & 0.708 \\
        6  & \texttt{MADLAD-400}              & Finetuning & Enc--dec & 0.707 \\
        7  & \texttt{Qwen3-8B}               & Few-shot   & Dec.\ & 0.690 \\
        8  & \texttt{Qwen3-0.6B}             & Finetuning & Dec.\ & 0.677 \\
        9  & \texttt{mT5-Base}               & Finetuning & Enc--dec & 0.660 \\
        10 & \texttt{mT5-Small}              & Finetuning & Enc--dec & 0.619 \\
        11 & \texttt{PhilTa}                 & Finetuning & Enc--dec & 0.599 \\
        12 & \texttt{MADLAD-400 $\rightarrow$ mT5-Small}   & Pipeline (TTS) & Enc--dec & 0.511 \\
        13 & \texttt{Qwen3-0.6B}             & Few-shot   & Dec.\ & 0.420 \\
        14 & \texttt{LatinSumm.\ $\rightarrow$ NLLB} & Pipeline (STT) & Enc--dec & 0.275 \\
        15 & \texttt{mT5-XLSum $\rightarrow$ NLLB}   & Pipeline (STT) & Enc--dec & 0.230 \\
        16 & \texttt{NLLB-200 $\rightarrow$ mT5-Small}     & Pipeline (TTS) & Enc--dec & 0.170 \\
        17 & \texttt{LongT5 $\rightarrow$ NLLB}      & Pipeline (STT) & Enc--dec & 0.136 \\
        \bottomrule
    \end{tabular}
    }
    \caption{Complete ranking of all 17 model--approach combinations by
    $S_{\text{Overall}}$.}
    \label{tab:cross_approach_ranking}
\end{table}
 
\begin{figure*}[ht]
    \centering
    \includegraphics[width=0.99\textwidth]{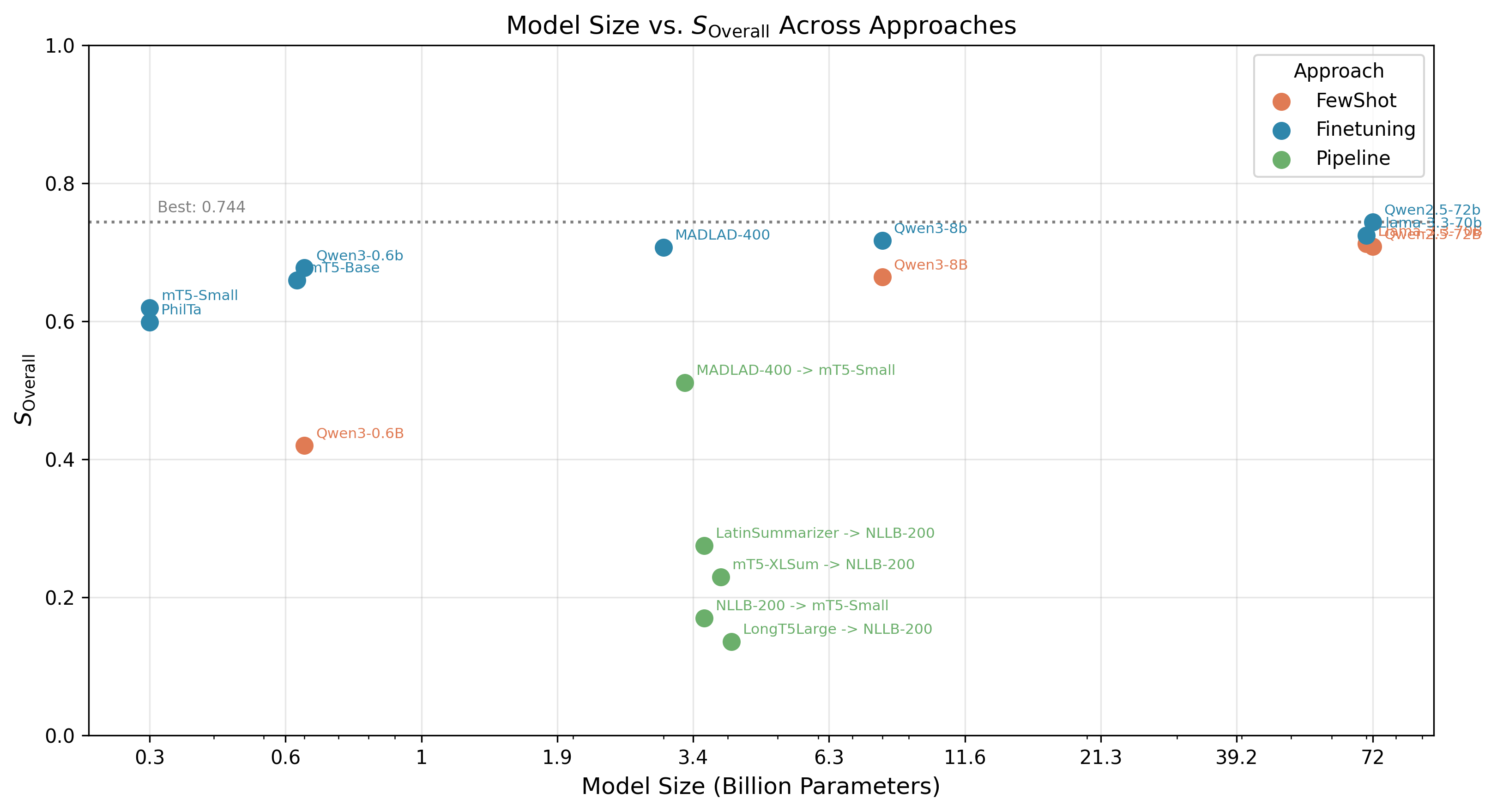}
    \caption{Parameters (log scale) vs.\ $S_{\text{Overall}}$ for all 17
    model--approach combinations. The dotted line marks the best score
    ($0.744$, finetuned \texttt{Qwen2.5-72B}).}
    \label{fig:params_vs_score}
\end{figure*}
 
At the category level, the approach winners differ most on the Lexical dimension, where finetuning ($0.332$) clearly beats few-shot ($0.233$), reflecting learned surface conventions of the training \textit{Kopfregesten}. On Constraints, however, few-shot prompting is essentially tied with finetuning ($0.846$ vs.\ $0.843$): explicit rule encoding in the prompt is nearly as effective as learning the conventions from data. The best models of both end-to-end paradigms are distributionally indistinguishable from the references ({MAUVE} $\geq 0.997$), while the pipeline lags on every axis (per-category radar in Appendix~\ref{a:radars}).

\paragraph{Parameter efficiency.}
Figure~\ref{fig:params_vs_score} plots parameters (log scale) against $S_{\text{Overall}}$. Finetuning achieves the best accuracy-per-parameter trade-off across the full 0.3B--72B range; few-shot is competitive only at $\geq$70B, collapsing to $0.420$ at 0.6B; pipelines cluster at 3--4B with scores $\leq 0.511$. Gains from scale flatten above $\sim$70B, where finetuned and few-shot models converge to within $\sim$0.03 points. The contrast between \texttt{MADLAD-400} finetuned end-to-end ($0.707$) and used as a pipeline stage ($0.511$) shows that approach design matters more than raw parameter count.
 
\paragraph{Few-shot vs.\ finetuning, model-controlled.} Because the same four decoder-only models were evaluated under both paradigms, Table~\ref{tab:fewshot_vs_finetune_delta} quantifies the finetuning gain per model. 

\begin{table}[ht]
    \centering
    \resizebox{.47\textwidth}{!}{
    \begin{tabular}{lrrr}
        \toprule
        \textbf{Model} & \textbf{Few-shot} & \textbf{Finetuned} & \textbf{$\Delta$ (abs.\ / rel.)} \\
        \midrule
        \texttt{Qwen3-0.6B}    & 0.420 & 0.677 & $+0.257$ ($+61.2\%$) \\
        \texttt{Qwen3-8B}      & 0.690 & 0.717 & $+0.027$ ($+3.9\%$)  \\
        \texttt{Qwen2.5-72B}   & 0.708 & 0.744 & $+0.036$ ($+5.1\%$)  \\
        \texttt{Llama-3.3-70B-Instruct} & 0.712 & 0.724 & $+0.012$ ($+1.7\%$)  \\
        \bottomrule
    \end{tabular}
    }
    \caption{Finetuning gain for the decoder-only models.}
    \label{tab:fewshot_vs_finetune_delta}
\end{table}
 
The gain is dominated by the smallest model: $+0.257$ ($+61.2\%$) for \texttt{Qwen3-0.6B}, but only $+1.7\%$ to $+5.1\%$ from 8B upward, with no ordering by size. For the smallest model, finetuning is transformative; for models that already perform well in-context, it is marginal.
 
\subsection{Discussion and Implications}
\label{subsec:discussion}
 
\paragraph{Finetuning vs.\ few-shot prompting.}
End-to-end finetuning is the most effective paradigm, yet its margin over few-shot prompting is surprisingly narrow at the top of the scale spectrum: the best finetuned model (\texttt{Qwen2.5-72B}, $0.744$) exceeds the best few-shot model (\texttt{Llama-3.3-70B-Instruct}, $0.712$) by only $0.032$ points---a gap that may narrow further with more training data or tuning. For practitioners with access to large instruction-following models but no training infrastructure, few-shot prompting with a guideline-explicit prompt is therefore a viable training-free alternative. The consequential difference lies at the small-model end: few-shot \texttt{Qwen3-0.6B} ($0.420$) is essentially unusable for \textit{Kopfregest} generation, whereas its finetuned counterpart ($0.677$, a $+61.2\%$ gain) becomes competitive. 
 
\paragraph{Encoder--decoder vs.\ decoder-only.} Among finetuned models, architectural family matters less than scale and multilingual pretraining breadth: the best encoder--decoder model (\texttt{MADLAD-400}, $0.707$) comes within $0.037$ of \texttt{Qwen2.5-72B} and surpasses \texttt{Qwen3-0.6B} ($0.677$). Pretraining breadth predicts performance more strongly than the encoder--decoder/decoder-only choice.

\paragraph{Why pipelines fail.} Pipeline errors are structural. First-stage errors propagate irreversibly because the second stage never attends to the original Latin: a single dropped entity (e.g., the recipient) invalidates the entire \textit{Kopfregest}. Compounding this, all components were used off-the-shelf, exposing the domain mismatch that few-shot models bridge via scale and finetuned models via adaptation.

\section{Conclusion}
\label{conclusion}

We investigated whether LLMs can automate the generation of German
\textit{Kopfregesten} from medieval Latin charters---a task that is
simultaneously low-resource, cross-lingual, and tightly
constrained by diplomatic convention. Our three research questions are
answered as follows.

\paragraph{Evaluation.} No single surface metric captures diplomatic
correctness. Our composite framework, which pairs lexical, semantic, and
distributional metrics with an LLM-as-a-Judge that evaluates six explicit
\textit{Kopfregest} rules, separates rule adherence from similarity to the references, but it does not verify whether the facts of a summary match the charter; this requires expert or reference-grounded fact-checking (see Limitations).

\paragraph{Architecture.} The theoretical advantages of encoder--decoder
models do not translate into practical gains in this domain. The best
encoder--decoder model (\texttt{MADLAD-400}, $0.707$) trails the best
decoder-only model (\texttt{Qwen2.5-72B}, $0.744$) yet remains competitive,
indicating that breadth of multilingual pretraining predicts performance more
strongly than the architectural distinction.

\paragraph{Training strategy.} End-to-end finetuning yields the best overall
quality and the best accuracy-per-parameter trade-off, but its margin over a
guideline-explicit few-shot prompt is narrow at scale ($0.032$ at $\sim$70B)
and decisive only for small models, transforming an unusable \texttt{Qwen3-0.6B}
($0.420$) into a competitive one ($0.677$, $+61.2\%$). A finetuned 8B model
($0.717$) already surpasses the best few-shot result of models with
$\sim$9$\times$ more parameters, while modular pipelines fail throughout
($\leq 0.511$) due to irreversible first-stage error propagation.

\paragraph{Practical takeaway.} Finetuned 3--8B models attain strong scores on
consumer hardware (Appendix~\ref{a:compute}), making them suitable as draft generators under expert review: since even strong systems still produce factual errors (Table~\ref{tab:e2e_examples}), they can support, but not replace, the editorial work. Practitioners with large
instruction-following models but no training infrastructure can instead adopt
guideline-explicit few-shot prompting as a viable production baseline. Future work includes reinforcement learning from verifiable diplomatic Constraints as rewards, domain-adaptive pretraining, and hybrid strategies that combine explicit directive prompting with finetuning.

\section*{Limitations}
Five caveats qualify these results. (i)~\emph{Overfitting}: with only
$1{,}332$ training pairs, training and validation loss diverge after
epoch~5, and the pattern is
expected across models; best-checkpoint selection by validation loss
mitigates but does not eliminate the risk of memorization rather than
generalization. (ii)~\emph{PEFT bias}: all runs used QLoRA, which restricts
adaptation to a low-rank subspace; if the domain shift required by medieval
Latin exceeds what low-rank updates can represent, the adapter becomes a
bottleneck, and full-parameter finetuning---infeasible under our compute
budget---might score higher. (iii)~\emph{Unified hyperparameters}: the recipe
was tuned via random search on \texttt{Qwen3-0.6B} only and transferred to
all models; this ensures a fair comparison under a fixed recipe but likely
leaves performance on the table for larger or architecturally different
models whose optimal learning rates, warmup schedules, or LoRA ranks differ.
(iv)~\emph{No expert evaluation}: our protocol is fully automatic, and the LLM
judge was not validated against human annotation; since it sees only the generated summary, it verifies the form of a \textit{Kopfregest} but not whether its content matches the charter. The only expert input we received was informal oral feedback on a small, non-randomly selected set of outputs, obtained without a fixed protocol, blinding, or independent annotators; we therefore do not report it as evidence.  Whether $S_{\text{Overall}}$ predicts factual correctness or the editorial effort saved per \textit{Kopfregest} is untested. A systematic expert evaluation of legal act, recipient, and object on a stratified sample of outputs is the most important next step.
(v)~\emph{Draft status}: our systems are draft generators, not a replacement
for editorial work, and have not yet been used in the editorial workflow.
Even the best system produces fluent, well-formed \textit{Kopfregesten} with
factual errors, such as a castle rendered as a city (Calatabiano;
Table~\ref{tab:e2e_examples}). Since such errors are easily missed when a
draft is read only for plausibility, every draft must be checked against the
Latin source. Our results thus support these models as an aid in an
expert-controlled workflow, not as autonomous generators of
\textit{Kopfregesten} for publication.

\section*{Ethics Statement}
We affirm that our research adheres to the \href{https://www.aclweb.org/portal/content/acl-code-ethics}{ACL Ethics Policy}. This work uses publicly available datasets and involves no human subjects or personally identifiable information. All code and data are released to enable reproducible research and further investigation.

\section*{Acknowledgments}

Esteban Garces Arias sincerely thanks the Mentoring Program of the Faculty of Mathematics, Statistics, and Informatics at LMU Munich and the Munich Center for Machine Learning (MCML) for their ongoing support.

\bibliography{custom}

\clearpage

\onecolumn

\section*{Appendix}

\appendix

\section{Dataset Details}
\label{a:dataset}
 
\subsection{Charter Entry Structure}
\label{a:entry-structure}

\begin{enumerate}[noitemsep]
    \item \textbf{Charter Number}: A unique sequential identifier (e.g., 1464)

    \item \textbf{Kopfregest}: A concise German summary describing the legal act,
        issuer, and recipient 

    \item \textbf{Date Line}: Standardized place and date of issue (e.g., ``Palermo,
        1199 Januar'')

    \item \textbf{Transmission}: Source documents (Originals, Copies), physical size,
        material, and archive location

    \item \textbf{Editions}: Bibliography of previous printed publications

    \item \textbf{Regesta/Literature}: References to other summaries and secondary
        literature

    \item \textbf{Commentary}: Editor's analysis covering paleography, chancery
        style, and historical context

    \item \textbf{Dorsal Notes}: Notes written on the back of the original parchment

    \item \textbf{Critical Apparatus}: Footnotes indicating textual variants or damage

    \item \textbf{Content Notes}: Numbered footnotes with historical explanations
\end{enumerate}

Only the Latin charter text (model input) and the \textit{Kopfregest}
(reference output) are extracted.
 
\subsection{Charter-to-Document Mapping}
\begin{table}[ht]
    \centering
    \begin{tabular}{lr}
        \toprule \textbf{Document} & \textbf{Charter Range} \\
        \midrule FII\_1            & 1--170                 \\
        FII\_2                     & 171--235               \\
        FII\_3                     & 236--299               \\
        FII\_4                     & 300--363               \\
        FII\_5                     & 364--426               \\
        FII\_6                     & 427--657               \\
        FII\_7                     & 658--929               \\
        FII\_8                     & 930--1199              \\
        FII\_9                     & 1200--1463             \\
        FII\_10                    & 1464--1721             \\
        \bottomrule
    \end{tabular}
    \caption{Charter-to-document mapping for the ten MGH Word documents
    (153--927 pages each).}
    \label{tab:doc-mapping}
\end{table}
 
\section{Extraction Pipeline Details}
\label{a:extraction}
 
\subsection{Regex-Based Extraction (Approach 1)}
The initial approach identified charter boundaries via the number pattern
\texttt{\^{}\textbackslash d\{1,4\}\textbackslash.}, used the place/date line
as a structural anchor, treated the first multi-line paragraph as the
condition/style section, extracted metadata blocks between anchors, collected
the remaining text as Latin content, and matched references with patterns such
as \texttt{(\textbackslash(SP|B\textbackslash).[...]\textbackslash))}. It
achieved 84\% successful extraction but failed on missing or reordered fields,
could not distinguish German summaries from German editorial annotations, and
required per-edge-case pattern maintenance.
 
\subsection{LLM-Based Semantic Extraction (Approach 2)}
Documents were exported to {HTML} via Google Docs (preserving, e.g.,
superscript footnote markers), parsed, segmented into per-charter blocks using
heading formatting cues, and automatically cleaned of editorial annotations:
all superscript elements (footnote and apparatus markers) were deleted, and
paragraphs containing several enumerated variant readings (e.g., ``a)~\ldots\
b)~\ldots'') were removed as critical-apparatus notes. The prompt additionally
instructed the model to exclude paragraphs that mix German and Latin, such as
scholarly annotations. The cleaned blocks were then passed to
Gemini~2.5~Flash, which returns {JSON} with keys \texttt{summary\_de} and
\texttt{latin\_text}.  An extraction counts as successful if the output is valid
{JSON} containing both keys; this was the case for 97\% of the charters.
Residual failures were limited to extremely long charters or corrupted {HTML}.
The extracted pairs were subsequently checked manually by members of the edition
project, who provided the data.
 
\paragraph{Extraction prompt (key elements).}
\begin{itemize}
    \item \textbf{Task}: ``Extract German summary and Latin text from a
        medieval charter.''
    \item \textbf{Fidelity}: ``Do NOT change a single word. Preserve ALL
        special characters and abbreviations.''
    \item \textbf{Boundaries}: the German summary ends immediately before the
        place/date line; the Latin text consists of consecutive purely-Latin
        paragraphs.
    \item \textbf{Edge cases}: normalize spaced names
        (``S.~A~n~g~e~l~o zu P~r~i~z~z~i'' $\rightarrow$ ``S.~Angelo zu
        Prizzi''); exclude status prefixes (``Verf\"alscht'',
        ``Zweifelhaft''); remove inline references.
    \item \textbf{Format}: ``Return ONLY valid {JSON} with keys
        \texttt{summary\_de} and \texttt{latin\_text}.''
\end{itemize}
 
\section{Few-Shot Prompting Details}
\label{a:fewshot}
 
\subsection{Prompt Versions}
\label{a:prompts}

\begin{figure*}[ht]
\begin{promptbox}{Version 1 (naive):}
\begin{Verbatim}[baselinestretch=0.92, breaklines=true, breaksymbolleft={}, breakindent=0pt]
                 
Du bist ein Experte für mittelalterliches Latein. Deine Aufgabe ist es,
lateinische Urkunden prägnant auf Deutsch zusammenzufassen. Die
Zusammenfassung sollte die wichtigsten Akteure, Orte und rechtlichen
Handlungen enthalten. Antworte NUR mit der deutschen Zusammenfassung.
 
\end{Verbatim}
\end{promptbox}
\begin{promptbox}{Version 2 (guideline-explicit):}
\begin{Verbatim}[baselinestretch=0.92, breaklines=true, breaksymbolleft={}, breakindent=0pt]
                 
Du bist ein Experte für mittelalterliche lateinische Urkunden.
Erstelle ein knappes Kopfregest auf Deutsch.
 
Regeln für das Kopfregest:
1. **Ein einziger Satz**: Das Regest darf niemals länger als ein Satz sein.
2. **Struktur**: [Name des Ausstellers] + [dispositives Verb] + [Empfänger]
   + [Rechtshandlung].
   - Beispiel: "Friedrich bestätigt dem Kloster Fulda alle Privilegien."
3. **Namen**: Nutze moderne deutsche Formen für Orte und Personen.
   Unbekannte Namen bleiben lateinisch.
4. **Inhalt**: Nur die rechtliche Verfügung. Keine Begründungen (Arenga),
   keine Strafandrohungen (Poenformel), keine Zeugen, keine Datierung.
5. **Stil**: Sachlich, präzise, keine Floskeln.
 
\end{Verbatim}
\end{promptbox}
\caption{Prompt Versions.}
\label{fig:prompts2}
\end{figure*}

\begin{figure}[!htbp]
    \centering
    \includegraphics[width=0.8\textwidth]{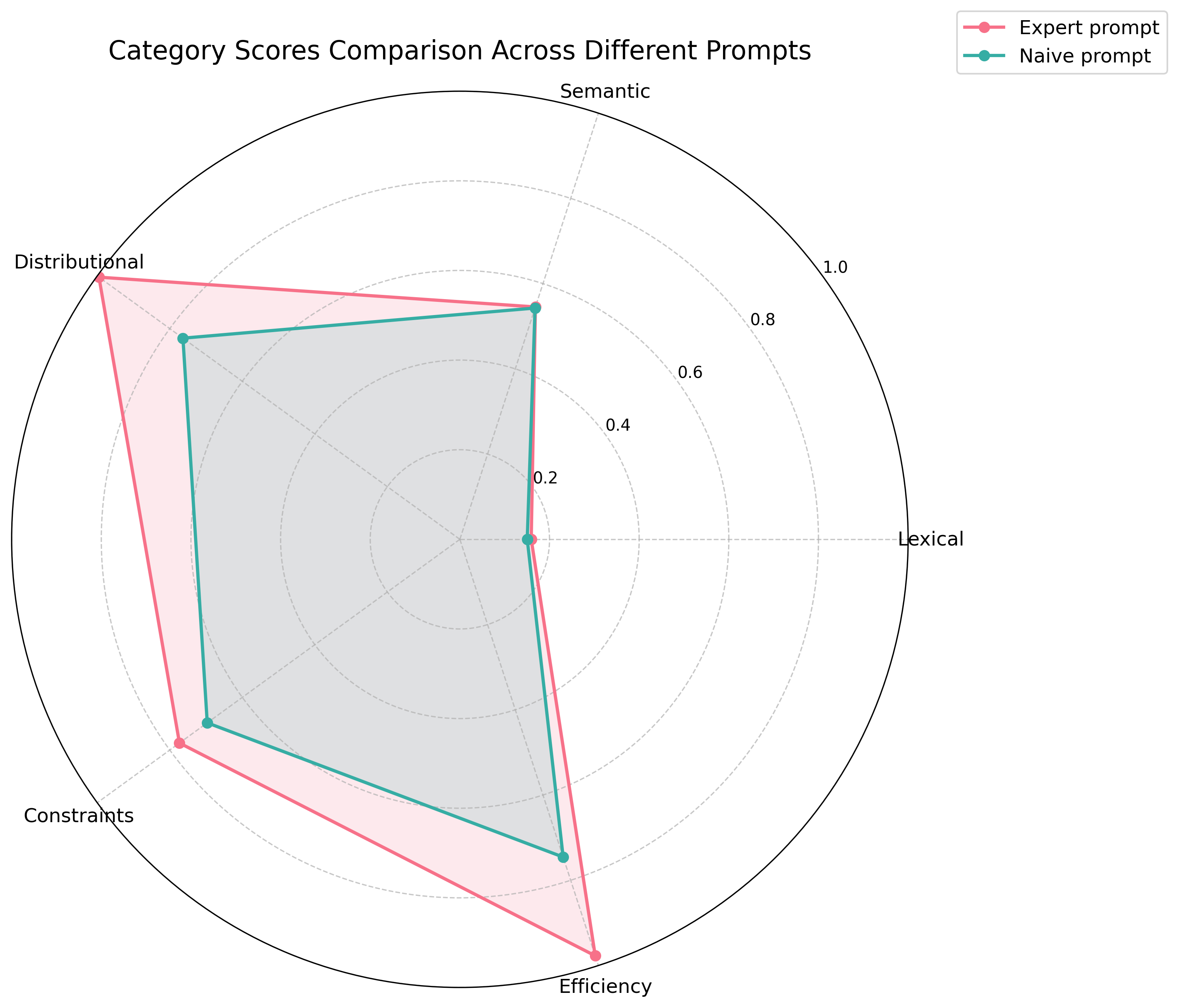}
    \caption{Naive vs.\ guideline-explicit prompt on \texttt{Qwen3-8B}
    ($n=10$ shots). Largest gains in Distributional and Efficiency
    ($+0.23$ each); Constraints $+0.08$; Lexical and Semantic nearly
    unchanged.}
    \label{fig:fewshot_prompt_radar}
\end{figure}
 
\subsection{Example Selection}
For each test sample, the $n$ demonstrations were drawn uniformly at random
from the training set without replacement (fixed seed 42), with no
similarity-based or stratified selection, isolating prompt and model effects
from retrieval quality and guaranteeing reproducibility.
 
\subsection{Tokenizer Behaviour on Medieval Latin}
\label{a:tokenizer}
\begin{table}[ht!]
\centering
\begin{tabular}{@{}lr@{}}
\toprule
                   & \textbf{Latin charters} \\
\midrule
Characters / token & 2.91 \\
Tokens / word      & 2.24 \\
Single-char tokens & 0.15 \\
\bottomrule
\end{tabular}
\caption{Tokenization statistics for the Latin inputs under the
\texttt{Qwen3-8B} tokenizer.}
\label{tab:tokenization}
\end{table}

Under the \texttt{Qwen3-8B} tokenizer, the Latin inputs fragment heavily
(Table~\ref{tab:tokenization}): the high tokens-per-word ratio and frequent
single-character fall-back indicate poor coverage of medieval orthography,
inflating effective sequence length and reducing the usable context available
for demonstrations and the charter itself.
 
\section{Pipeline Details}
\label{a:pipeline}

\subsection{Configuration Details}
\label{a:pipeline_config}

All pipeline components are encoder--decoder models; generation uses beam search (4 beams, no sampling) for deterministic output.
 
\paragraph{Translation step (TTS).}
The two translators control the target language by different mechanisms.
\texttt{NLLB-200} uses dedicated language-code tokens
(\texttt{src\_lang}$=$\texttt{lat\_Latn}, \texttt{forced\_bos\_token\_id}$=$%
\texttt{deu\_Latn}), so source and target are fixed entirely at the tokenizer and
decoder level without any text-level prefix. \texttt{MADLAD-400} instead prepends
the text tag \texttt{<2de>} (``translate to German'') to the input.
 
\paragraph{Summarization step.}
\texttt{mT5-Small} (finetuned on German MLSum) and all three {STT} Latin
summarizers (\texttt{LatinSummarizer}, \texttt{LongT5-Large}, \texttt{mT5-XLSum})
follow the standard T5 multi-task paradigm, prepending the prefix
\texttt{"summarize:~"} with no further instructions; the model conditions solely on this prefix.
 
\paragraph{Translation step (STT).}
\texttt{NLLB-200} translates the Latin abstract to German using the same language-code mechanism as above.

\subsection{Full Pipeline Results}
\label{a:pipeline_full}
Table~\ref{tab:pipeline_scores} gives the per-category breakdown behind the
overall scores in the main text. The category profile confirms that failures are
concentrated in the Lexical and Distributional dimensions---where first-stage
errors are most visible---while the comparatively high Efficiency scores of some
{STT} configurations merely reflect outputs of plausible length that are nonetheless semantically empty.

\begin{table}[!htbp]
    \centering
    \resizebox{.9\textwidth}{!}{
    \begin{tabular}{llrrrrrr}
        \toprule \textbf{Dir.} & \textbf{Configuration} & $S_{\text{Lex}}$ & $S_{\text{Sem}}$ & $S_{\text{Dist}}$ & $S_{\text{Cons}}$ & $S_{\text{Eff}}$ & $S_{\text{Overall}}$ \\
        \midrule
        TTS & \texttt{MADLAD-400 $\rightarrow$ mT5-Small}     & \textbf{0.169} & \textbf{0.496} & \textbf{0.715} & 0.363 & \textbf{0.813} & \underline{\textbf{0.511}} \\
        TTS & \texttt{NLLB-200 $\rightarrow$ mT5-Small}       & 0.002 & 0.279 & 0.016 & 0.290 & 0.263 & 0.170 \\
        \midrule
        STT & \texttt{LatinSummarizer $\rightarrow$ NLLB-200} & 0.007 & 0.316 & 0.019 & 0.283 & \textbf{0.749} & 0.275 \\
        STT & \texttt{mT5-XLSum $\rightarrow$ NLLB-200}       & 0.018 & 0.177 & 0.009 & 0.193 & 0.751 & 0.230 \\
        STT & \texttt{LongT5-Large $\rightarrow$ NLLB-200}    & 0.002 & 0.278 & 0.009 & 0.323 & 0.069 & 0.136 \\
        \bottomrule
    \end{tabular}
    }
    \caption{Full pipeline results for both directions. No configuration
    approaches the few-shot baseline. Best score per column in bold.}
    \label{tab:pipeline_scores}
\end{table}
 
\section{Finetuning Details}
\label{a:finetune}
 
\subsection{Adapter Placement}
Decoder-only models: {LoRA} adapters on all four attention projections
(query, key, value, output). Encoder--decoder models: adapters on query and
value projections only.

\subsection{Encoder--Decoder Model Overview}
\label{a:encdec_models}
Table~\ref{tab:encdec_models} summarizes the four finetuned encoder--decoder
models. All share the T5 architecture with gated-GELU feed-forward layers and
differ mainly in size, vocabulary, and pretraining data. For \texttt{MADLAD-400}
we use the 3B machine-translation checkpoint (\texttt{google/madlad400-3b-mt}).
The larger 7B and 10B variants were not used. The same checkpoint serves as the
translator in the {TTS} pipeline (Section~\ref{subsec:pipeline-strategies}).

\begin{table}[!htbp]
    \centering
    \resizebox{\textwidth}{!}{
    \begin{tabular}{llrcrrrrlr}
        \toprule
        \textbf{Model} & \textbf{Checkpoint} & \textbf{Params} & \textbf{Layers (enc/dec)} & $d_{\text{model}}$ & $d_{\text{ff}}$ & \textbf{Heads} & \textbf{Vocab} & \textbf{Pretraining data} & \textbf{Max.\ input} \\
        \midrule
        \texttt{MADLAD-400}~\citep{Kudugunta2023} & \texttt{google/madlad400-3b-mt} & 2.94B & 32/32 & 1{,}024 & 8{,}192 & 16 & 256{,}000 & MADLAD-400 + parallel MT data (450+ languages) & 1{,}024 \\
        \texttt{mT5-Base}~\citep{Xue2021}         & \texttt{google/mt5-base}        & 0.58B & 12/12 & 768     & 2{,}048 & 12 & 250{,}112 & mC4 (101 languages)                            & 512 \\
        \texttt{mT5-Small}~\citep{Xue2021}        & \texttt{google/mt5-small}       & 0.30B & 8/8   & 512     & 1{,}024 & 6  & 250{,}112 & mC4 (101 languages)                            & 512 \\
        \texttt{PhilTa}~\citep{Riemenschneider2023} & \texttt{bowphs/PhilTa}        & 0.30B & 12/12 & 768     & 2{,}048 & 12 & 64{,}103  & Ancient Greek, Latin, and English texts         & 1{,}024 \\
        \bottomrule
    \end{tabular}
    }
    \caption{Finetuned encoder--decoder models, sorted by overall score as in
    Table~\ref{tab:e2e_results_all}. Architecture values are taken from the
    released Hugging Face configurations. Parameter counts for \texttt{MADLAD-400}
    and \texttt{PhilTa} come from the released checkpoints; those for \texttt{mT5}
    follow \citet{Xue2021}. \emph{Max.\ input} is the source truncation length
    (tokens) used during finetuning; the target length was capped at 256 tokens
    for all models.}
    \label{tab:encdec_models}
\end{table}
 
\subsection{Hyperparameter Search}
\label{a:hpo}
Hyperparameter optimization was run only on the smallest decoder-only model
(\texttt{Qwen3-0.6B}): it is cheap enough for many trials yet still exhibits the
core difficulties of the domain (long inputs, domain mismatch, strict output
style). All trials used 10 epochs for comparability, and model selection used
$S_{\text{Overall}}$ as the objective, since it integrates lexical/semantic
similarity, distributional match, constraint compliance, and conciseness into a
single criterion. The random search covered: learning rate \{1e-4, 1.5e-4, 2e-4,
2.5e-4, 3e-4\}; warmup steps \{10, 20, 30, 40, 50, 60, 70\}; weight decay \{0.01, 0.02,
0.05, 0.1\}; LoRA rank $r$ \{8, 16, 32\}; batch size \{2, 4\}; gradient
accumulation \{2, 4, 8\}; label smoothing \{0.0, 0.05, 0.1, 0.15\}; and max
gradient norm \{0.3, 0.5, 1.0\}. The best configuration
(Table~\ref{tab:e2e_best_hparams}) was then transferred unchanged to all
remaining models---a pragmatic compromise that avoids giving any subset of models
an unfair tuning advantage and isolates architectural and scale effects under a
fixed recipe.
 
\begin{table}[!htbp]
    \centering
    \begin{tabular}{ll}
        \toprule \textbf{Hyperparameter} & \textbf{Value}    \\
        \midrule Epochs                  & 10                \\
        Batch size                       & 2                 \\
        Gradient accumulation steps      & 4                 \\
        Effective batch size             & 8                 \\
        Learning rate                    & 2.5e-4            \\
        Warmup steps                     & 10                \\
        Weight decay                     & 0.1               \\
        Max grad norm                    & 0.5               \\
        Label smoothing                  & 0.1               \\
        LoRA rank $r$                    & 16                \\
        LoRA alpha                       & 32 ($=2\times r$) \\
        LoRA dropout                     & 0.05              \\
        \bottomrule
    \end{tabular}
    \caption{Best hyperparameter configuration from random search on
    \texttt{Qwen3-0.6B}.}
    \label{tab:e2e_best_hparams}
\end{table}
 
\begin{figure}[!htbp]
    \centering
    \includegraphics[width=0.5\textwidth]{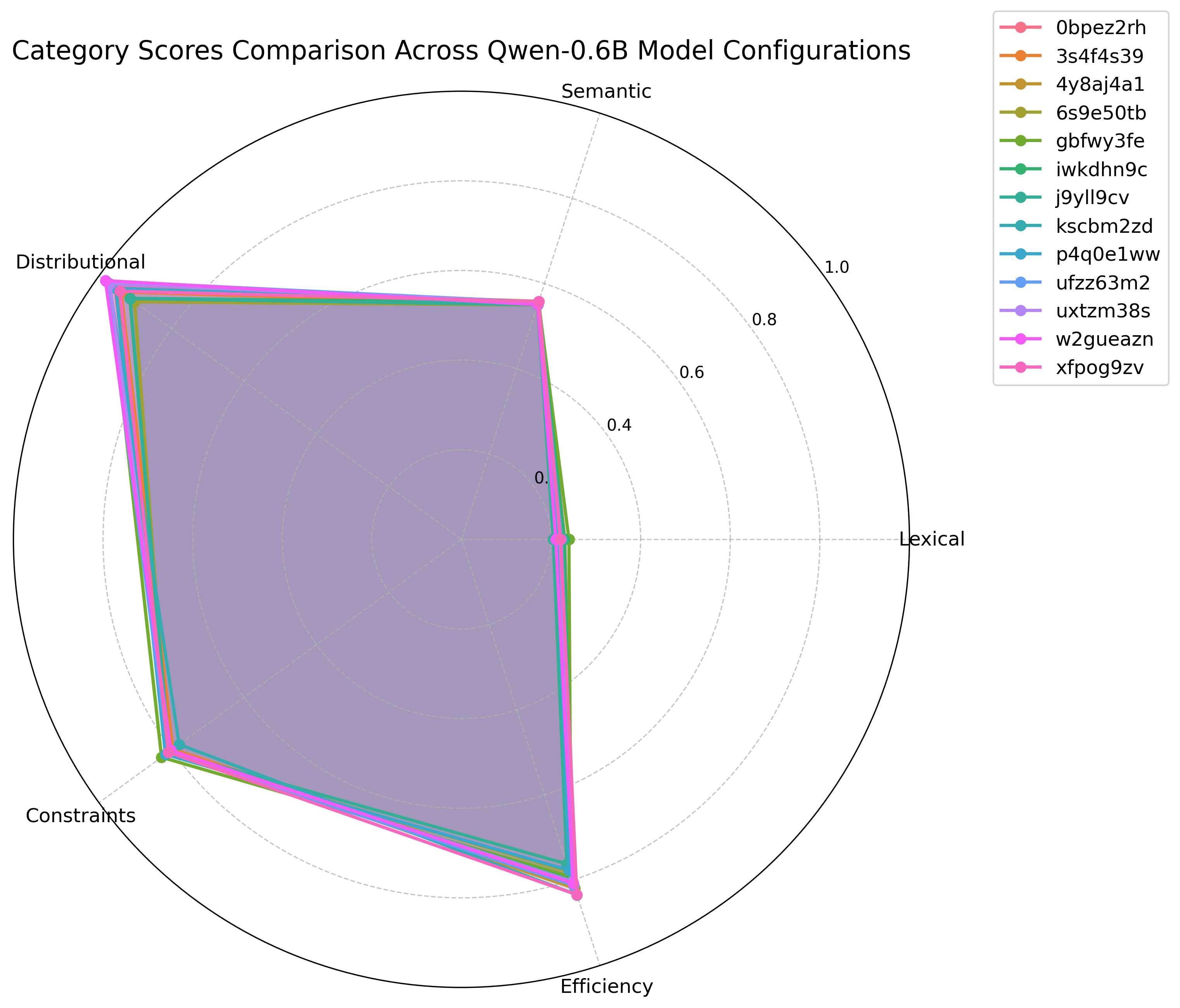}
    \caption{$S_{\text{Overall}}$ for all random-search runs on
    \texttt{Qwen3-0.6B}; best run highlighted. The small spread
    ($\Delta S_{\text{Overall}}=0.03$) indicates robustness to the searched
    hyperparameter range.}
    \label{fig:e2e_hpo_runs}
\end{figure}
 
\subsection{Training Dynamics}
\label{a:loss}
\begin{figure}[!htbp]
    \centering
    \includegraphics[width=0.95\textwidth]{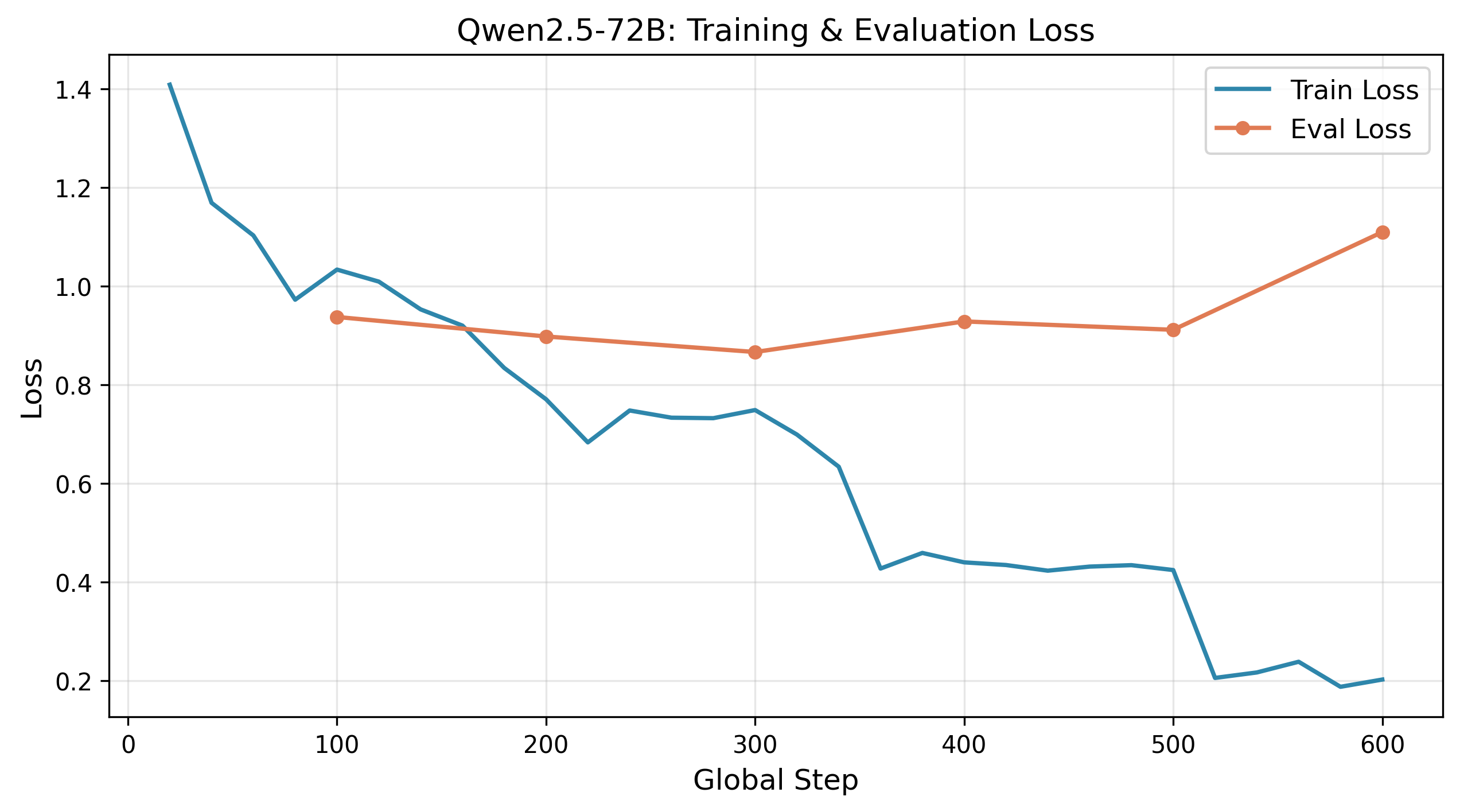}
    \caption{Training and validation loss for \texttt{Qwen2.5-72B} over 10
    epochs (600 steps). Training loss falls monotonically ($1.41 \rightarrow
    0.20$); validation loss reaches its minimum of $0.866$ at step~300
    (epoch~5) and rises to $1.110$ thereafter, indicating overfitting. The
    step-300 checkpoint was used for all test-set predictions.}
    \label{fig:e2e_loss_best}
\end{figure}

We tracked training and validation loss for all runs to detect overfitting.
Figure~\ref{fig:e2e_loss_best} shows the curves for the best model
(\texttt{Qwen2.5-72B}). Training loss falls monotonically ($1.41\rightarrow0.20$),
while validation loss reaches its minimum ($0.866$) at step~300 (epoch~5) and
rises thereafter to $1.110$---a clear divergence indicating that the model begins
to memorize training examples rather than generalize. The step-300 checkpoint,
selected by minimum validation loss, was used for all test-set predictions; the
same qualitative pattern is expected across models given the small dataset.

\section{Computing Infrastructure}
\label{a:compute}

All experiments were run on an HPC cluster with SLURM
scheduling and one GPU per job; no data or model parallelism was
used. Models up to 70B/72B parameters were trained on NVIDIA A100 GPUs with
80~GB of memory. Decoder-only models were finetuned with QLoRA in 4-bit NF4 precision (double quantization,
float16 compute dtype); \texttt{MADLAD-400}, \texttt{mT5}, and
\texttt{PhilTa} were finetuned in bfloat16 with LoRA adapters and
gradient checkpointing. No experiment was executed on consumer hardware; the
deployment claim in Section~\ref{conclusion} is an extrapolation
from the measured memory footprints, not a direct measurement.

\section{Per-Family Category Scores}
\label{a:radars}
\begin{figure}[!htbp]
    \centering
    \includegraphics[width=0.49\textwidth]{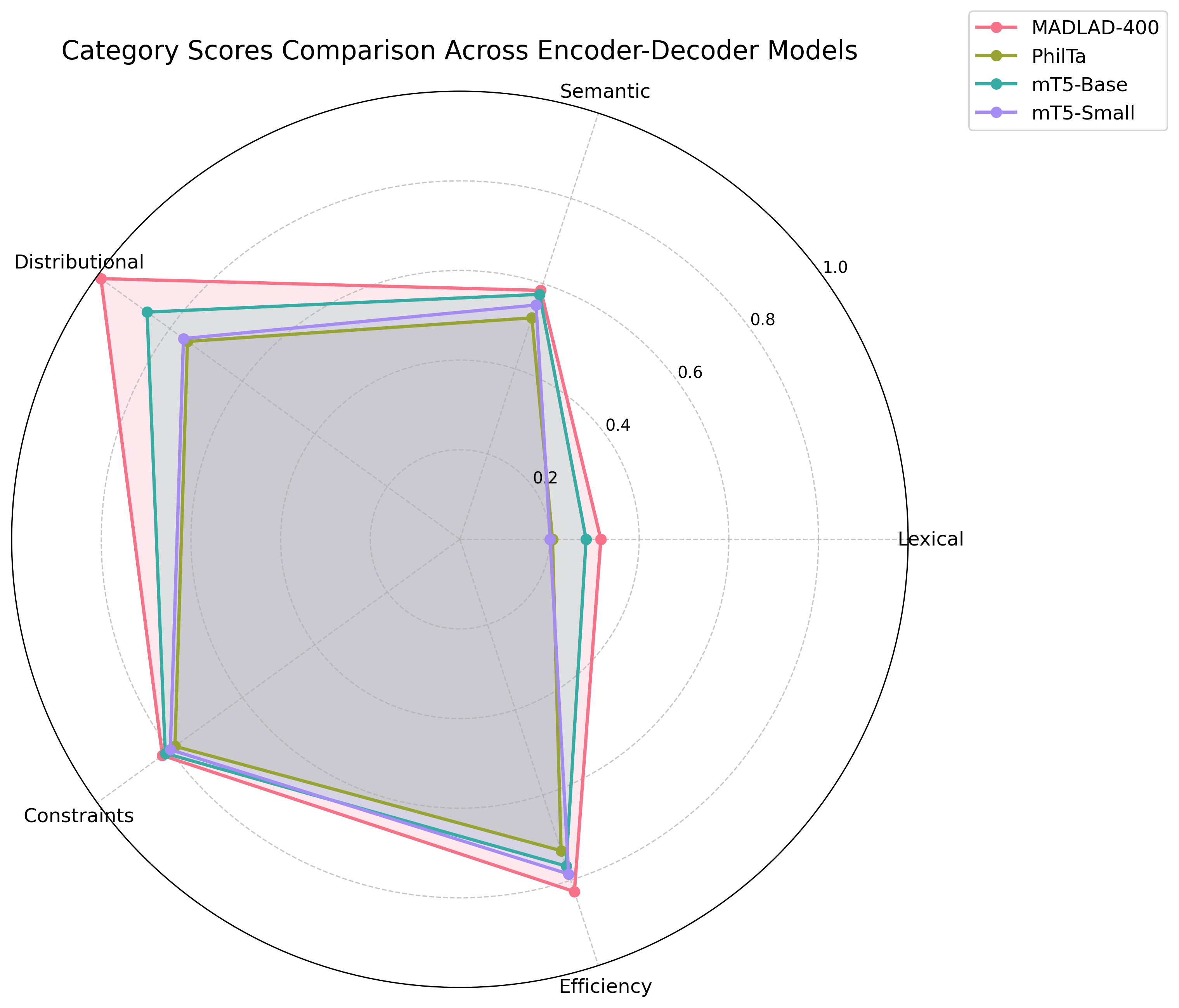}
    \includegraphics[width=0.49\textwidth]{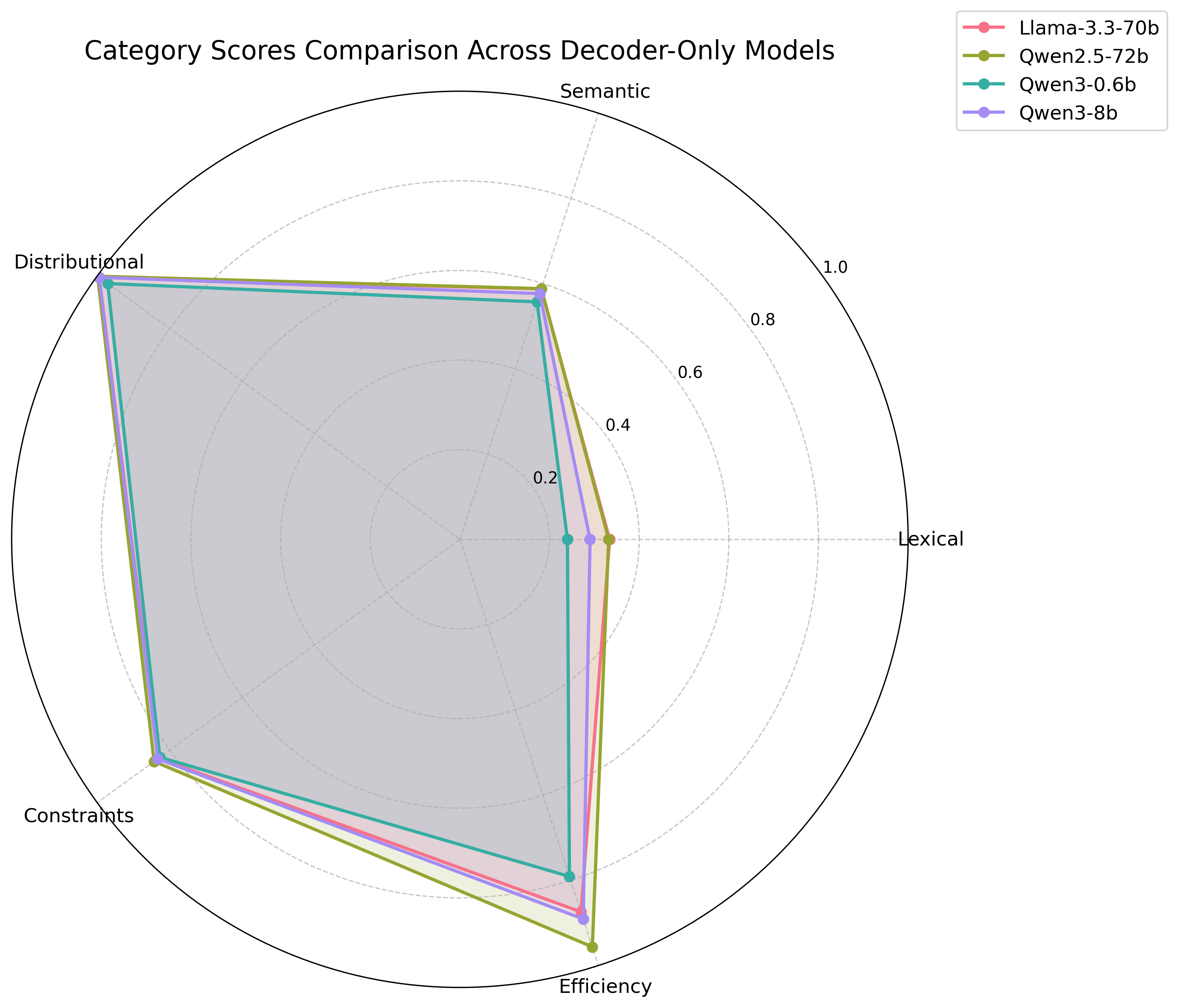}
    \caption{Category scores for finetuned encoder--decoder models (left;
    \texttt{MADLAD-400} best, $S_{\text{Overall}}=0.707$) and decoder-only
    models (right; \texttt{Qwen2.5-72B} best, $S_{\text{Overall}}=0.744$).}
    \label{fig:e2e_radars}
\end{figure}

\begin{figure}[!htbp]
    \centering
    \includegraphics[width=0.49\textwidth]{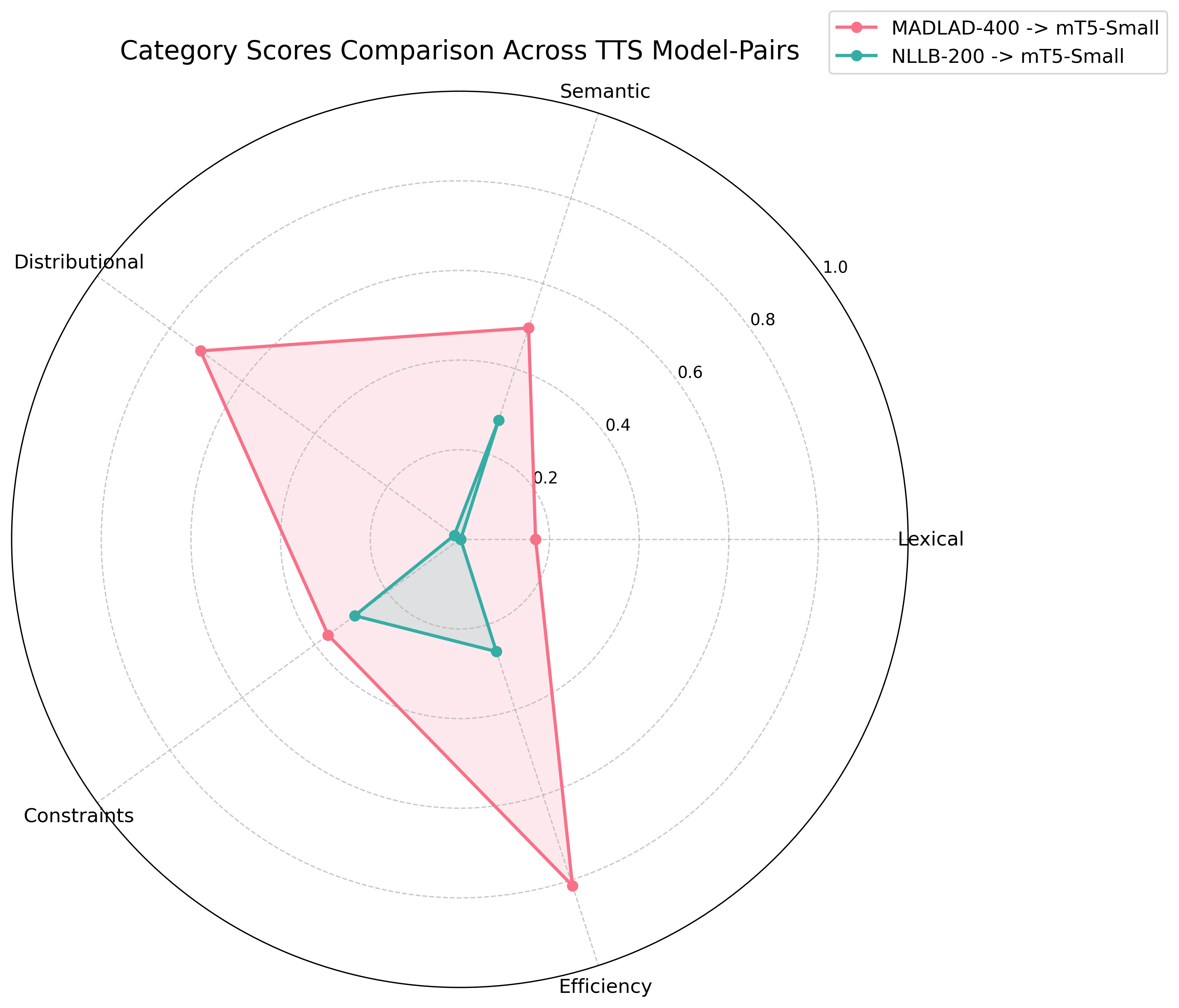}
    \includegraphics[width=0.49\textwidth]{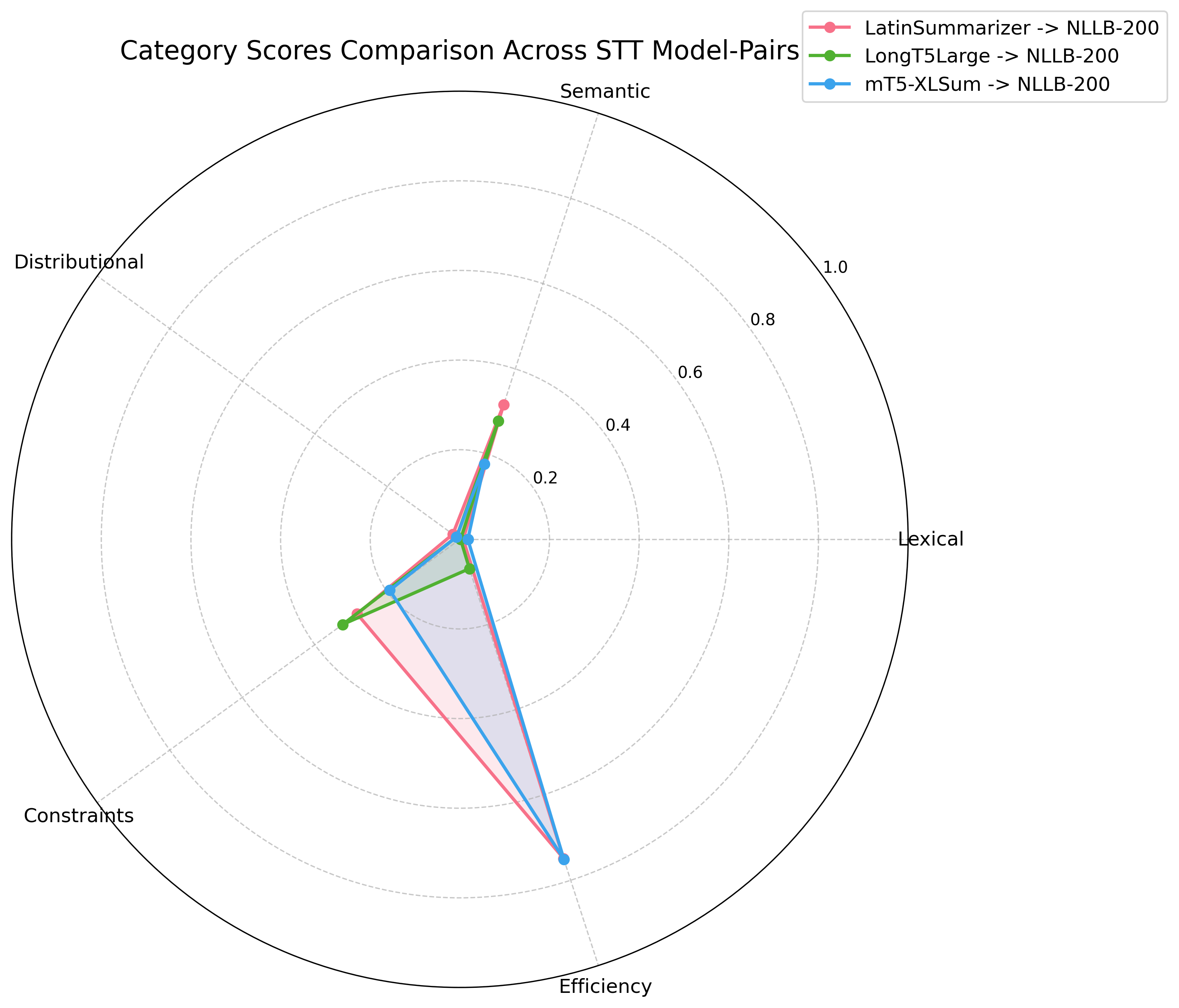}
    \caption{Category scores for {TTS} (left) and {STT} (right) pipeline
    configurations. \texttt{NLLB-200}-based {TTS} fails almost completely
    ($S_{\text{Overall}}=0.170$); all {STT} configurations fail
    ($S_{\text{Overall}}<0.28$) due to first-stage error propagation.}
    \label{fig:pipeline_radars}
\end{figure}

The following radar plots decompose the aggregate scores into the five categories,
making visible where each model family gains or loses. Among finetuned models
(Figure~\ref{fig:e2e_radars}), the decoder-only family dominates on Distributional
and Efficiency, while the encoder--decoder gap is driven mainly by Lexical.
The pipeline radars (Figure~\ref{fig:pipeline_radars}) show the opposite profile: collapse concentrated in Lexical and Distributional, consistent with first-stage error propagation.

\begin{figure}[!htbp]
    \centering
    \includegraphics[width=0.5\textwidth]{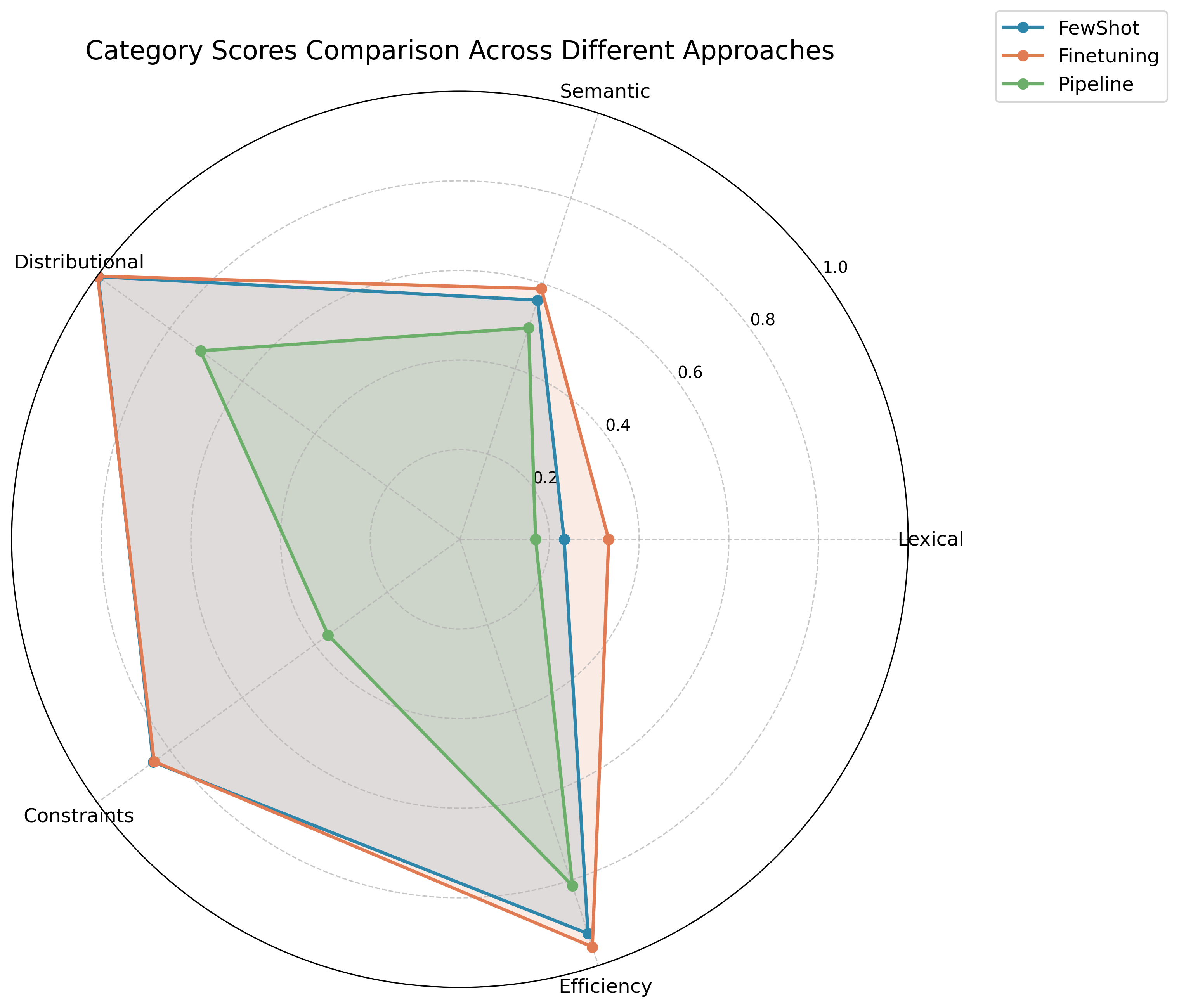}
    \caption{Category scores for the best model of each approach:
    finetuned \texttt{Qwen2.5-72B} ($0.744$),
    few-shot \texttt{Llama-3.3-70B-Instruct} ($0.712$), and
    \texttt{MADLAD-400~$\rightarrow$~mT5-Small} pipeline ($0.511$).}
    \label{fig:cross_approach_category_radar}
\end{figure}

\section{Sensitivity to Category Weights}
\label{a:weights}
To test whether our conclusions depend on equal weighting, we recompute
$S_{\text{Overall}}$ for all 17 model--approach combinations under alternative
weightings: (i)~leaving out one category at a time, (ii)~weighting all 14
metrics equally instead of the five categories, and (iii)~10{,}000 weight
vectors drawn uniformly from the simplex (Dirichlet$(1,\dots,1)$, fixed seed).
We compare each resulting ranking with the reported one (Table~\ref{tab:cross_approach_ranking}) using Kendall's $\tau$ and check
whether the main claims of Section~\ref{sec:results} still hold
(Table~\ref{tab:weight_sensitivity}).

\begin{table}[!htbp]
    \centering
    \resizebox{.85\textwidth}{!}{
    \begin{tabular}{lrrl}
        \toprule
        \textbf{Weighting} & \textbf{Kendall's $\tau$} & \textbf{Max.\ rank shift} & \textbf{Top model} \\
        \midrule
        Without Lexical               & 0.90 & 4 & \texttt{Qwen2.5-72B} (FT) \\
        Without Semantic              & 1.00 & 0 & \texttt{Qwen2.5-72B} (FT) \\
        Without Distributional        & 0.96 & 1 & \texttt{Qwen2.5-72B} (FT) \\
        Without Constraints           & 0.99 & 1 & \texttt{Qwen2.5-72B} (FT) \\
        Without Efficiency            & 0.88 & 3 & \texttt{Qwen2.5-72B} (FT) \\
        Equal weights per metric (14) & 0.90 & 2 & \texttt{Qwen2.5-72B} (FT) \\
        \midrule
        Random weights ($n{=}10{,}000$) & 0.91 (mean); 0.82 (5th pct.) & 2 (median) & \texttt{Qwen2.5-72B} (FT) in 99.1\% \\
        \bottomrule
    \end{tabular}
    }
    \caption{Rank stability of the 17 model--approach combinations under
    alternative weightings of the composite score, relative to the equally
    weighted categories used in the paper.}
    \label{tab:weight_sensitivity}
\end{table}

The ranking is stable across all variants. The following claims hold under
every fixed variant and under at least 94\% of the random weightings.
Finetuned \texttt{Qwen2.5-72B} is the best model (99.1\%). The best finetuned
model outperforms the best few-shot model (99.2\%). The finetuning gain is
largest for \texttt{Qwen3-0.6B} (99.5\%). The best pipeline ranks below all
end-to-end models except few-shot \texttt{Qwen3-0.6B} (94.1\%) and finetuned
\texttt{MADLAD-400} outperforms its pipeline counterpart (100\%). Two claims
depend on the weighting: finetuned \texttt{Qwen3-8B} outperforms the best
few-shot model in only 62\% of the random weightings, and finetuning improves
over few-shot prompting for all four decoder-only models in only 69\%. Both
fail when the Lexical category is removed, i.e., they rest on the finetuned
models' advantage in adopting the surface wording of the reference
\textit{Kopfregesten}. This is consistent with the per-category analysis in
Section~\ref{subsec:performance-comparison}.

\section{Qualitative Examples}
\label{a:examples}

\begin{table}[!htbp]
    \centering
    \small
    \begin{tabular}{p{0.28\textwidth}p{0.32\textwidth}p{0.32\textwidth}}
        \toprule \textbf{Latin input (truncated)}                                                                                                                                                                                                                                                                    & \textbf{Prediction}                                                                                                                                                                                                                                                                                                                                                   & \textbf{Reference}                                                                                                                                                                                                                                                                                                                                                                                                                                                          \\
        \midrule \textit{In nomine sancte et individue trinitatis. Fridericus secundus divina favente clementia Romanorum imperator semper augustus et rex Sicilie. Inter alia pietatis opera, quibus imperialis dignitas uberius exornatur, liberalitas obtinet principatum \ldots}                                 & Friedrich nimmt den Templerorden unter seinen besonderen Schutz, bestätigt ihm alle von seinen Vorfahren, insbesondere von Friedrich~I. und Heinrich~VI., verliehenen Privilegien, Freiheiten und Immunitäten und droht jedem, der den Orden verletzt, seine Ungnade und eine Geldstrafe an.                                                                          & Friedrich nimmt den Meister und die Brüder des Templerordens mit all ihren Gütern im ganzen Kaiserreich in seinen besonderen Schutz und bestätigt ihnen und ihren Nachfolgern alle Privilegien seines Großvaters Friedrich~I., seines Vaters Heinrich~VI. sowie anderer Herrscher.                                                                                                                                                                                          \\
        \midrule \textit{In nomine sancte et individue trinitatis. Fridericus secundus divina favente clementia Romanorum imperator semper augustus, Ierusalem et Sicilie rex. Si inter curas regiminis temporalis, cuius monarchiam divina clementia miserante tenemus, ecclesiarum profectibus consulentes \ldots} & Friedrich gestattet dem Abt des Zisterzienserklosters S.~Salvatore al Monte Amiata auf Bitten des Papstes Gregor und des Abtes Rainald die Umwandlung des Benediktinerklosters S.~Salvatore al Monte Amiata in ein Zisterzienserkloster und bestätigt ihm alle Besitzungen und Rechte sowie die Schutz- und Schirmbriefe seines Vaters Heinrich und seiner Vorfahren. & Friedrich gestattet auf Bitten des Abtes Rainer des Benediktinerklosters S.~Salvatore di Monte Amiata und auf die Forderung Papst Gregors~IX. hin, das unter den schwarzen Mönchen in spiritueller und materieller Hinsicht heruntergekommene Kloster in eines des Zisterzienserordens umzuwandeln, nimmt Abt und Kloster in seinen besonderen Schutz und bestätigt ihnen die Privilegien seiner Vorfahren, alles vorbehaltlich der ihm und dem Reich geschuldeten Dienste. \\
        \bottomrule
    \end{tabular}
    \caption{Two qualitative examples for the best few-shot model (\texttt{Llama-3.3-70B-Instruct}).
    The Latin inputs are truncated for better readability.}
    \label{tab:fewshot_examples}
\end{table}

\begin{table}[!htbp]
    \centering
    \small
    \begin{tabular}{p{0.19\textwidth}p{0.22\textwidth}p{0.18\textwidth}p{0.28\textwidth}}
        \toprule
        \textbf{Latin input (truncated)} & \textbf{1. Prediction: Translation (truncated)} & \textbf{2. Prediction: Kopfregest} & \textbf{Reference} \\
        \midrule
        \textit{In nomine sancte et individue trinitatis. Fridericus secundus divina
        favente clementia Romanorum imperator semper augustus et rex Sicilie. Inter
        alia pietatis opera, quibus imperialis dignitas uberius exornatur, liberalitas
        obtinet principatum \ldots}
        &
        Im Namen der heiligen und individuellen Dreifaltigkeit. Friedrich der
        Zweite, der göttliche Gnadenkönig der römischen Kaiser, immer Augustus und
        König von Sizilien. \ldots\ erwirbt er die Freizügigkeit des Fürstentums \ldots
        &
        Friedrich der Zweite, der göttliche Gnadenkönig der römischen Kaiser, immer
        Augustus und König von Sizilien, erwirbt er die Freizügigkeit des Fürstentums.
        &
        Friedrich nimmt den Meister und die Brüder des Templerordens mit all ihren
        Gütern im ganzen Kaiserreich in seinen besonderen Schutz und bestätigt ihnen
        und ihren Nachfolgern alle Privilegien seines Großvaters Friedrich~I., seines
        Vaters Heinrich~VI. sowie anderer Herrscher. \\
        \midrule
        \textit{In nomine sancte et individue trinitatis. Fridericus secundus divina
        favente clementia Romanorum imperator semper augustus, Ierusalem et Sicilie
        rex. Si inter curas regiminis temporalis, cuius monarchiam divina clementia
        miserante tenemus, ecclesiarum profectibus consulentes \ldots}
        &
        (C.)\ Im Namen der heiligen und individuellen Dreifaltigkeit. \ldots\ Wenn
        unter den Pferden des zeitlichen Regimes, dessen Monarchie wir die göttliche
        Gnade erbarmungslos halten \ldots
        &
        Wenn unter den Pferden des zeitlichen Regimes, dessen Monarchie wir die
        göttliche Gnade erbarmungslos halten, wollen wir die Bitten erfüllen.
        &
        Friedrich gestattet auf Bitten des Abtes Rainer des Benediktinerklosters
        S.~Salvatore di Monte Amiata und auf die Forderung Papst Gregors~IX.\ hin,
        das unter den schwarzen Mönchen in spiritueller und materieller Hinsicht
        heruntergekommene Kloster in eines des Zisterzienserordens umzuwandeln,
        nimmt Abt und Kloster in seinen besonderen Schutz und bestätigt ihnen die
        Privilegien seiner Vorfahren, alles vorbehaltlich der ihm und dem Reich
        geschuldeten Dienste. \\
        \bottomrule
    \end{tabular}
    \caption{Two qualitative examples for the best-performing {TTS} configuration
    (\texttt{MADLAD-400} $\rightarrow$ \texttt{mT5-Small}), illustrating semantic
    error propagation from translation (Step~1) to the final summary (Step~2).
    The Latin inputs are truncated for readability. In Example~1, the translation
    entirely omits the Templar content and instead produces a paraphrase of the
    opening formula. In Example~2, \texttt{MADLAD-400} mistranslates \textit{curas}
    (duties/cares) as \textit{Pferde} (horses), and the summarizer faithfully
    inherits this.}
    \label{tab:pipeline_tts_examples}
\end{table}

\begin{table}[!htbp]
    \centering
    \small
    \begin{tabular}{p{0.28\textwidth}p{0.32\textwidth}p{0.32\textwidth}}
        \toprule \textbf{Latin input (truncated)} & \textbf{Prediction} & \textbf{Reference} \\
        \midrule \textit{In nomine sancte et individue trinitatis amen. Ego Fredericus dei gratia Romanorum rex semper augustus et rex Sycilie omnibus in perpetuum. Licet dilecti fideles nostri principes fidelitate obligati sint maiestati nostre regie, decens tamen et honestum fore reputamus \ldots} & Friedrich restituiert seinem Verwandten Bischof Konrad von Regensburg alle Rechte, die seine Kirche von seinen Vorgängern und ihm selbst erhalten hat, sowie alle guten Gewohnheiten, die die Bürger der Stadt auf irgendeine Weise in seine Gewalt gebracht haben, und bestätigt ihm und seiner Kirche alle bisher gefundenen und zukünftig noch zu findenden Salzquellen sowie Gold-, Silber-, Blei-, Kupfer- und andere Metallvorkommen. & Friedrich anerkennt und restituiert Bischof Konrad~IV. von Regensburg alle Rechte und guten Gewohnheiten in der Stadt, welche die Regensburger Bischöfe jemals vom Reich hatten, aber auf Betreiben der Bürger zurückgenommen worden sind, und überträgt der Regensburger Kirche alle Salz-, Erz- sowie sonstigen Metallvorkommen, die man auf ihren Gütern gefunden hat und noch finden wird. \\
        \midrule \textit{Fredericus divina favente clementia rex Sicilie, ducatus Apulie et principatus Capue. Etsi divine maiestatis intuitu, que nostri regiminis gloriam cum felicitate multiplicat, personas ecclesiasticas et loca venerabilia pia volumus affectione diligere \ldots} & Friedrich verleiht dem Erzbischof Berard und der Kirche von Messina in Anbetracht der Verdienste um seine Eltern und ihn selbst die Stadt Calatabiano mit allem Zubehör. & Friedrich überläßt dem Erzbischof Berard und der Kirche von Messina in Anbetracht der Treue und der Dienste zugunsten seiner Eltern sowie der Mühen und Aufwendungen, die er und seine Kirche für ihn selbst erbracht haben, das Kastell Calatabiano. \\
        \bottomrule
    \end{tabular}
    \caption{Two qualitative examples for the best end-to-end finetuned model
    (\texttt{Qwen2.5-72B}). The Latin inputs are truncated for better readability.
    Example~1 shows correct structure but over-specifies metal categories not
    enumerated in the reference. Example~2 misidentifies Calatabiano as a city
    (\textit{Stadt}) rather than a castle (\textit{Kastell}) and uses \textit{verleiht}
    instead of the more precise \textit{überläßt}.}
    \label{tab:e2e_examples}
\end{table}

\twocolumn

\end{document}